\documentclass[twoside,american]{elsarticle}
\usepackage[T1]{fontenc}
\usepackage[latin9]{inputenc}
\usepackage{amsmath}
\usepackage{amsthm}
\usepackage{amssymb}
\makeatletter
\ifx\proof\undefined
\newenvironment{proof}[1][\protect\proofname]{\par
	\normalfont\topsep6\p@\@plus6\p@\relax
	\trivlist
	\itemindent\parindent
	\item[\hskip\labelsep\scshape #1]\ignorespaces
}{%
	\endtrivlist\@endpefalse
}
\providecommand{\proofname}{Proof}
\fi

\makeatletter
\def\ps@pprintTitle{%
 \let\@oddhead\@empty
 \let\@evenhead\@empty
 \let\@oddfoot\@empty
 \let\@evenfoot\@empty}
\makeatother

\makeatother

\theoremstyle{plain}
\newtheorem{thm}{\protect\theoremname}
\theoremstyle{definition}
\newtheorem{defn}[thm]{\protect\definitionname}
\newtheorem{example}[thm]{\protect\examplename}
\theoremstyle{plain}
\newtheorem{cor}[thm]{\protect\corollaryname}
\newtheorem{lem}[thm]{\protect\lemmaname}
\newtheorem{prop}[thm]{\protect\propositionname}
\usepackage{babel}
\providecommand{\corollaryname}{Corollary}
\providecommand{\definitionname}{Definition}
\providecommand{\examplename}{Example}
\providecommand{\lemmaname}{Lemma}
\providecommand{\propositionname}{Proposition}
\providecommand{\theoremname}{Theorem}

\begin{document}
\begin{frontmatter}
\title{Mean-based incomplete pairwise comparisons method with the reference
values}
\author[weaiib]{Konrad~Ku\l akowski\corref{cor1}}
\ead{kkulak@agh.edu.pl}
\author[wms]{Anna~K\k{e}dzior}
\ead{kedzior@wms.mat.agh.edu.pl}
\author[wms]{Jacek Szybowski}
\ead{szybowsk@agh.edu.pl}
\author[opa]{Jiri Mazurek}
\ead{szybowsk@agh.edu.pl}
\cortext[cor1]{Corresponding author}
\address[weaiib]{AGH University of Science and Technology, Faculty of Electrical Engineering,
Automatics, Computer Science and Biomedical Engineering, Kraków, Poland}
\address[wms]{AGH University of Science and Technology, Faculty of Applied Mathematics,
Kraków, Poland}
\address[opa]{Department of Informatics and Mathematics, Silesian University in
Opava, School of Business Administration in Karvina, Czech Republic}
\begin{abstract}
In this article, we propose two quantitative methods for calculating
weight vectors for incomplete pairwise comparison matrices using reference
values. Both procedures are extensions of arithmetic and geometric
heuristic estimation (HRE) methods. The proposed solutions allow flexible
selection of the number of reference alternatives and the range of
comparisons, from the acceptable minimum to a complete set. In this
paper, we prove that the newly introduced geometric HRE method for
incomplete data is optimal. For this method, we also prove the existence
of a feasible solution. In the paper, we also provide sufficient conditions
for the existence of a solution for the second arithmetic variant
of the HRE method. We illustrate the presented methods with numerical
examples.
\end{abstract}
\end{frontmatter}

\section{Introduction}

Decision-making based on pairwise comparisons of alternatives has
a long history. Although the first well-documented use of the pairwise
comparison method is attributed to Llull \citep{Colomer2011rlfa},
we may suspect that the idea itself is much older. The ancient Egyptian
Osiris judgment ceremony \citep{Carelli2011tbod}, whose essence was
to compare the deeds of the deceased (symbolized by the heart) against
the appropriate values (signified by an ostrich feather - the sign
of the goddess Maat) on a scale, may be a clue. Llull and his successors,
such as Cusanus, Condorcet, and Copeland \citep{Condorcet1785eota,Brandt2016hocs,McLean1995cosc}
used qualitative pairwise comparisons. It means that they did not
specify the strength of preferences, but only their nature: who is
the winner and who is the loser. The quantitative use of pairwise
comparisons can be traced to the twentieth-century researcher Thurstone,
who developed a method for describing social values \citep{Thurstone1927tmop}.
Perhaps the best-known example of the use of quantitative pairwise
comparison for decision-making is the Analytic Hierarchy Process (AHP)
method proposed by Saaty \citep{Saaty1977asmf} in the second half
of the twentieth century. In addition to AHP, the idea of comparing
alternatives in pairs can be found in many other decision-making methods,
including BWM \citep{Brunelli2019ambw,Rezaei2015bwmc}, ELECTRE, PROMETHEE
\citep{Greco2016mcda}, MACBETH \citep{BanaECosta2005otmf}, and techniques
such as multiobjective optimization \citep{Greco2010iemo} or multiple
criteria sorting \citep{Kadzinski2015mabp}. Despite its long history,
the pairwise-comparisons method is still the subject of research and
development. Starting from considering the original AHP method \citep{Dyer1990rota,BanaeCosta2008acao,Koczkodaj2016pcrs}
through examining the inconsistency of a set of comparisons \citep{Brunelli2018aaoi,Aguaron2020ttgc,Mazurek2019snpo,Kulakowski2018iito},
uncertainty handling \citep{Mohammadi2020bbwm,Benitez2019moup,Ramik2018sras},
incomplete PC matrix processing \citep{kulakowski2020otgm,Bozoki2010ooco,Agoston2022itfi,Faramondi2019iahp},
and others. A broader overview of the topics related to the pairwise
comparison method and AHP can be found in the books \citep{Kulakowski2020utahp,Ramik2020pcmt}. 

Among decision-making methods, there is a group based on reference
values. Within this group, Helwig\textquoteright s method \citep{Roszkowska2024aceo}
is worth mentioning. It does not limit the number of reference objects
(here called diagnostic variables). When building recommendations,
the relative closeness of the reference object and the development
pattern is calculated. The TOPSIS method \citep{Panda2018timc} uses
two reference points: the best and worst alternatives. The reference
objects are combinations of each criterion's best and worst possible
values. Similar to the TOPSIS, the VIKOR (VIseKriterijumska Optimizacija
I Kompromisno Resenje) method \citep{Opricovic2004csbm} defines reference
values, though the calculation process for the recommendation differs.
The idea of best- and worst-reference points can also be found in
the BWM (Best-Worst Method) \citep{Rezaei2015bwmc}. This method can
be seen as a combination of optimization ideas derived from pairwise
comparison methods and distance-based techniques, such as TOPSIS.
The use of reference values in combination with the pairwise comparison
method can be found in the HRE (Heuristic Rating Estimation) method
\citep{Kulakowski2014hrea,kulakowski2014hreg,Kulakowski2015note},
an extension of which is described in this paper, as well as in Parsimonious
AHP (PAHP) \citep{Abastante2019anpa}. In the latter, the authors
define a five-step decision process in which the decision maker (DM)
first assigns each alternative a rating relative to a given criterion,
then selects a set of reference alternatives, which are subsequently
subjected to a pairwise comparison procedure. In the fourth step of
the proposed process, the DM can reconsider the results and possibly
modify the selected values to achieve a satisfactory level of inconsistency.
The fifth step of the procedure is to calculate the final weight values
for non-reference alternatives using linear interpolation. In the
context of the pairwise comparison method, it is also worth mentioning
AHP-express \citep{Leal2020aeas}, in which an arbitrarily selected
best alternative serves as the reference element. It is worth noting
that AHP-express is similar to the Pairwise Comparisons Simplified
\citep{Koczkodaj2015pcs}, supplemented with an indication of a reference
alternative. In addition, the reference objects are used in the BIPOLAR
method \citep{KonarzewskaGubala2002mccb}, DARP (Distances to Aspiration
Reference Points) method \citep{Roszkowska2020uiac}, or COMET method
\citep{Salabun2014tcom}. 

The heuristic rating estimation (HRE) method, the extension of which
is the subject of this paper, is one of the decision-making techniques
based on pairwise comparisons \citep{Kulakowski2014hrea}. At the
core of HRE is the assumption that the priorities of some (reference)
alternatives are initially known. Hence, weights need to be calculated
only for non-reference alternatives with previously unknown ranking
values. Initially, we defined arithmetic HRE assuming that the priority
of a single alternative should correspond to the arithmetic mean of
the comparisons-weighted priorities of other options \citep{Kulakowski2014hrea,Kulakowski2015ahre,Kulakowski2016note}.
This assumption makes the HRE similar to the Eigenvalue method (EVM)
\citep{Saaty1977asmf} for calculating priorities in AHP \citep{Kulakowski2016srot}.
Later, we defined geometric HRE, which has assumptions somewhat similar
to those of the Geometric Mean Method (GMM) \citep{Kulakowski2015hreg,Crawford1987tgmp}.
Both arithmetic and geometric HRE, however, assume the existence of
all possible pairwise comparisons, i.e., between any two of the alternatives
considered. In the case of AHP, EVM, and GMM methods, the assumption
that a complete set of comparisons must be used has been relaxed \citep{Harker1987amoq,kulakowski2020otgm}. 

Therefore, in this paper, we propose procedures to calculate the weight
vector for an incomplete set of comparisons using both arithmetic
and geometric HRE. Thanks to the proposed solution, we gain the freedom
to select both the number of reference alternatives and the comparisons
to be made when preparing the weight vector. It allows for $n$ alternatives
to freely determine the number of pairwise comparisons between alternatives
from a minimum of $n-1$ to a complete set of $n(n-1)/2$ comparisons.
Offering two variants (arithmetic and geometric) of HRE allows DM
to choose the procedure for setting priorities. In our work, we show
that the newly defined geometric incomplete HRE (giHRE) inherits the
optimality of the GMM method, i.e., the calculated weight vector minimizes
the least squares error function. We also present theorems showing
that, in practice, a solution always exists for giHRE. Those who prefer
the arithmetic mean over the geometric approach can use the arithmetic
incomplete HRE (aiHRE). In this case, we present two statements defining
sufficient (though not necessary) conditions for the existence of
a solution. We also define aiHRE as an optimization problem that can
be helpful to some extent where algebraic solutions fail. We have
illustrated the presented solutions with numerical examples. The advantage
of the presented giHRE and aiHRE methods is their easy integration
with the AHP hierarchical model. This means that each node in the
hierarchical model can, but need not, be replaced with such a method.
It gives DM the freedom to decide in which context a given alternative
is a reference alternative. Integration with the hierarchical model
allows, in particular, for the identification of reference criteria.
An essential feature of the HRE approach is the ability to expand
the model with additional alternatives while maintaining the weights
of the reference alternatives. It offers protection against the undesirable
phenomenon of rank reversal \citep{Belton1983oasc}. The similarity
to EVM and GMM methods, and the fact that the starting point for model
construction is a complete or incomplete pairwise comparison matrix,
enables the use of the standard concepts of pairwise comparison matrix
inconsistency \citep{Brunelli2018aaoi,Kulakowski2020iifi}.

The paper is organized as follows: Section \ref{sec:Preliminaries}
contains basic information about the quantitative method of pairwise
comparison, including EVM and GMM procedures. In addition, it recalls
key concepts from algebra and matrix theory. Section \ref{sec:Heuristic-Rating-Estimation}
briefly presents the arithmetic and geometric HRE method \citep{Kulakowski2014hrea,Kulakowski2015ahre}
for complete matrices. Section \ref{sec:Additive-Incomplete-Heuristic}
defines the aiHRE method. The introduced procedure is illustrated
with a numerical example. Section \ref{sec:Multiplicative-Incomplete-Heuris}
defines the giHRE. As before, the solution is illustrated with a numerical
example. Section \ref{sec:Solution-of-Incomplete} contains two theorems
on sufficient conditions for the existence of a solution for the aiHRE
method, a definition of aiHRE as an optimization problem, two theorems
on the existence of a solution for the giHRE method, and a theorem
on the optimality of the giHRE method. Two sections conclude the entire
work, \ref{sec:Discussion} and \ref{sec:Summary}, containing a discussion
and a summary. 

\section{Preliminaries\label{sec:Preliminaries}}

\subsection{Pairwise comparisons}

The pairwise comparisons (PC) method is the process designed to transform
the set of comparisons into a ranking of alternatives. Let $A=\{a_{1},\ldots,a_{n}\}$
denote a set of alternatives while $C=[c_{ij}]$ means a set of comparisons
in the form of $n\times n$ matrix, where $c_{ij}\in\mathbb{R}_{+}$
for $i,j=1,\ldots,n$. Each $c_{ij}$ implies the result of a direct
comparison between $a_{i}$ and $a_{j}$. When the result of the given
comparison $c_{ij}$ is unknown, we write that $c_{ij}=c_{ji}=?$. 
\begin{defn}
A PC matrix $C$ is said to be reciprocal if $c_{ij}=1/c_{ji}$ for
all $i,j=1,\ldots,n$ except when $c_{ij}=?$.
\end{defn}
The purpose of the PC method is to calculate the weight vector.
\begin{defn}
Let the weight function for $A$ be $w:A\rightarrow\mathbb{R}_{+}$,
such that if $w(a_{i})>w(a_{j})$ for some $0<i,j\leq n$, then $a_{i}$
is more preferred than $a_{j}.$ 
\end{defn}
The above function prioritizes the various alternatives. The more
preferred alternatives have higher weights. The function $w$ is usually
represented as a priority vector in the form:
\begin{equation}
w=\left(w(a_{1}),w(a_{2}),\ldots,w(a_{n})\right)^{T}.\label{eq:prior-vect}
\end{equation}
The fact that $c_{ij}$ corresponds to the ratio of the preferential
strength of alternatives $a_{i}$ and $a_{j}$ implies that one may
expect that $c_{ij}=w(a_{i})/w(a_{j})$. What, in turn, leads to the
postulate of transitivity\footnote{As $c_{ik}=w(a_{i})/w(a_{k})$ and $c_{kj}=w(a_{k})/w(a_{j})$.},
i.e., $c_{ij}=c_{ik}c_{kj}$ for every triad $i,j,k=1,\ldots,n$.
Unfortunately, because the vector $w$ is computed based on all pairwise
comparisons, in practice, we may expect only that $c_{ij}\approx w(a_{i})/w(a_{j})$,
thus $c_{ij}\approx c_{ik}c_{kj}$. 
\begin{defn}
An $n\times n$ PC matrix $C$ is said to be inconsistent if there
exist $i,j,k=1,\ldots,n$ such that $c_{ij}\neq c_{ik}c_{kj}$.

It is easy to prove that if $c_{ij}=c_{ik}c_{kj}$ then $c_{ij}=w(a_{i})/w(a_{j})$
\citep[p. 96]{Kulakowski2020utahp}, regardless of the prioritization
method, provided, of course, that $c_{ij}$ means the ratio between
the preferential strength of $a_{i}$ and $a_{j}$. 

There are at least a dozen methods for computing a vector $w$ \citep{Choo2004acff,Kulakowski2020utahp}.
The two most popular are the Eigenvalue Method (EVM) and the Geometric
Mean Method (GMM). Saaty originally proposed the first one in his
seminal paper \citep{Saaty1977asmf} which is based on the concept
of a vector and the eigenvalue of the matrix $C$. So, let $C=[c_{ij}]$
be a PC matrix containing expert judgments for $n$ alternatives,
and $w_{\textit{max}}$ be a principal eigenvector of $C$, i.e. 
\[
Cw_{\textit{max}}=\lambda_{\textit{max}}w_{\textit{max}},
\]
where $\lambda_{\textit{max}}$ is a principal eigenvalue (spectral
radius) of $C$. Thus, the priority vector $w_{\textit{ev}}$ (\ref{eq:prior-vect})
is a rescaled version of $w_{\textit{max}}$, i.e. 
\[
w_{\textit{ev}}=\frac{1}{\sum^{n}_{i=1}w_{\textit{max}}(a_{i})}w_{\textit{max}}.
\]
The second method, although it is based on similar premises \citep{Kulakowski2016srot},
is easier to calculate. In GMM, the priority of the i-th alternative
is the appropriately rescaled geometric mean of all its direct comparisons.
I.e. 
\begin{equation}
w_{\textit{gm}}(a_{i})=\alpha\left(\prod^{n}_{j=1}c_{ij}\right)^{1/n},\label{eq:GMM-eq}
\end{equation}

where 
\[
\alpha=\left(\sum^{n}_{i=1}\left(\prod^{n}_{j=1}c_{ij}\right)^{1/n}\right)^{-1}.
\]
Both the EVM and GMM methods have incomplete counterparts. Extending
the EVM to incomplete matrices was proposed by Harker \citep{Harker1987amoq}.
A similar extension of GMM for incomplete PC matrices can be found
in \citep{kulakowski2020otgm}, or, equivalently, for the Logarithmic
Least-Squares method in \citep{Bozoki2010ooco,Tone1993llsm}. 
\end{defn}
\begin{example}
As an illustrative example, consider buying a large refrigerator for
our home. Our selection is limited because we are looking for an uncommon
$70$-cm-wide model. Eventually, we find three suitable refrigerators
offered at similar prices. To choose the best one, we compare the
refrigerators pairwise and record the results in a matrix $C$. For
a pair $i\!-\!j$, we assign a score of $1$ if we like them equally,
$2$ if we slightly prefer refrigerator $i$ to refrigerator $j$,
and $3$ if we moderately prefer refrigerator $i$ to refrigerator $j$.
If $c_{ij}=p$, where $p\in\{1,2,3\}$, then we set $c_{ji}=1/p$.
Of course, $c_{ii}=1$. The resulting matrix is shown below:
\[
C=\left[\begin{array}{ccc}
1 & 3 & \frac{1}{2}\\
\frac{1}{3} & 1 & \frac{1}{2}\\
2 & 2 & 1
\end{array}\right].
\]
As a result, for both the EVM and GMM methods, we obtain: 
\[
w_{ev}=w_{\text{gm}}=\left[0.348,0.167,0.483\right]^{T}.
\]
Thus, regardless of the method used, refrigerator No. $3$ is the
best option.
\end{example}

\subsection{Matrices and functions}

In the rest of this paper, we will use selected concepts from matrix
theory. For convenience, some of them are listed below.
\begin{thm}
\citep[p. 20]{Varga2000mia} An $n\times n$ matrix $C$ is irreducible
if and only if its directed graph $G(C)$ is strongly connected.
\end{thm}
The above statement has a natural interpretation in the context of
the pairwise comparison method. Namely, if the alternatives represented
by the rows and columns of the matrix $C$ are the nodes of the graph,
and comparisons $c_{ij}$ define the edge (if they exist), it is irreducible
if every two alternatives are at least indirectly comparable to each
other. The latter means that we can create a sequence of comparisons
for any two alternatives $a_{i}$ and $a_{j}$ such that $c_{i,p_{1}},c_{p_{1},p_{2}},\ldots,c_{p_{q},j}$,
and each of the comparisons exists.
\begin{defn}
\label{def:diag-dom-matrix}\citep[p. 22]{Varga2000mia} An $n\times n$
complex matrix $C=[c_{ij}]$ is diagonally dominant if 
\begin{equation}
\left|c_{ii}\right|\geq\sum^{n}_{\stackrel{j=1}{j\neq i}}\left|c_{ij}\right|,\label{eq:diagnoally-dominance-def}
\end{equation}
for all $1\leq i\leq n$. An $n\times n$ matrix $C$ is strictly
diagonally dominant if strict inequality in (\ref{eq:diagnoally-dominance-def})
is valid for all $1\leq i\leq n$. 
\end{defn}
Extending the above definition to include the concept of irreducibility,
we have
\begin{defn}
\label{def:irreducibly-dom-matrix}\citep[p. 22]{Varga2000mia} A
matrix $C$ is irreducibly diagonally dominant if $C$ is irreducible
and diagonally dominant with strict inequality holding in (\ref{eq:diagnoally-dominance-def})
for at least one $i$. 
\end{defn}
The above two definitions create the conditions for making the following
statement. 
\begin{thm}
\label{thm:varga-theorem}\citep[p. 23]{Varga2000mia} Let $C=[c_{ij}]$
be an $n\times n$ strictly or irreducibly diagonally dominant complex
matrix. Then the matrix $C$ is nonsingular. If all the diagonal entries
of $C$ are in $\mathbb{R}_{+}$, then for all the eigenvalues $\lambda_{i}$
of $C$, for $i=1,\ldots,n$, holds $\text{Re(\ensuremath{\lambda}}_{i})>0$.
\end{thm}
The above statement leads to the observation that every square Hermitian
matrix with a strongly dominant diagonal or a matrix with an irreducibly
dominant diagonal, where additionally the diagonal is real and positive,
is positively definite \citep[p. 23]{Varga2000mia}. Some diagonally
dominant matrices can be M-matrices. Consider the following theorem
\citep[p. 90]{Varga2000mia}.
\begin{thm}
\label{thm:invert-matrix-exists}If $C=[c_{ij}]$ is a real $n\times n$
matrix with $c_{ij}\leq0$ for all $i\neq j$, then the following
are equivalent:
\begin{itemize}
\item $C$ is nonsingular, and $C^{-1}>0$.
\item The diagonal entries of $C$ are positive real numbers. If $D$ is
a diagonal matrix such that $d_{ii}=1/c_{ii}$, then the matrix $B=I-DA$
is nonnegative, irreducible and convergent. 
\end{itemize}
\end{thm}
Based on this statement, the following observation can be formulated
and proven \citep[p. 91]{Varga2000mia}.
\begin{cor}
\label{cor:postivie-matrix-observ} If $C=[c_{ij}]$ is a real, irreducibly
diagonally dominant $n\times n$ matrix with $c_{ij}\leq0$ for all
$i\neq j$, and $a_{ii}>0$, for all $i=1,\ldots,n$, then $A^{-1}>0$.
\end{cor}
Following Plemmons \citep[p. 176]{Plemmons1976mcnm}, let us adopt
the classical definition of an m-matrix.
\begin{defn}
An $n\times n$ matrix $C$ in the form $C=sI-B$, where $B\geq0$,
and the spectral radius $\rho$ such that $s\geq\rho(B)$ is called
an M-matrix.
\end{defn}
In particular, Plemmons proves that an inverse-positive matrix, i.e.,
a matrix $C$ for which $C^{-1}$ exists and $C^{-1}\geq0$, is an
M-matrix. Based on Plemmons' theorem \citep[p. 180]{Plemmons1976mcnm},
we can say the following. 
\begin{cor}
\label{cor:m-matrix-corollary} The matrix that satisfies the conditions
of Corollary \ref{cor:postivie-matrix-observ} is an M-matrix.
\end{cor}
 In this paper, we also make a simple observation about the form
of inverse matrices. Formulated as a lemma, it looks as follows.
\begin{lem}
\label{lem:eth-row-of-inverse}If matrix $C=[c_{ij}]$ is invertible
and its k-th row $C^{T}_{k}\overset{\text{df}}{=}\left[c_{k1},\dots,c_{kn}\right]^{T}$
has the form $e^{T}_{k}\overset{\text{df}}{=}\left[0,\dots,0,\underbrace{1}_{k\text{-th position}},0,\dots,0\right]^{T}$,
i.e. $C^{T}_{k}=e^{T}_{k}$, then the k-th row of the inverse matrix
$C^{-1}$ is identical to the k-th row of $C$.
\end{lem}
\begin{proof}
Since it is $CC^{-1}=I$ then we know that $C_{k}\cdot C^{T}_{k}=\sum_{i}c^{2}_{ki}=1.$
Because $c_{kk}=1$ then $\sum_{i,i\neq k}c^{2}_{ki}=0$, what is
possible only if $c_{k1}=\ldots=c_{k,k-1}=c_{k,k+1}=c_{k,n}=0$. 
\end{proof}

\section{Heuristic Rating Estimation Method\label{sec:Heuristic-Rating-Estimation}}

The central assumption of the HRE method is that there are two subsets
of alternatives. The first $A_{U}$, consisting of alternatives with
unknown preferential values, and the second $A_{K}$, consisting of
reference alternatives for which the preferences are known. We assume,
without loss of generality, that $A_{U}=\{a_{1},\ldots,a_{k}\}$ and
$A_{K}=\{a_{k+1},\ldots,a_{n}\}$. Thus, the set of alternatives is
given as $A=A_{U}\cup A_{K}$ where $A_{U}\cap A_{K}=\varnothing$.
For $a_{r}\in A_{K}$ the ranking value $w(a_{r})\in\mathbb{R}_{+}$
is known and fixed at the beginning of the decision process. It follows
that comparisons $c_{ij}$ where $i,j=k+1,\ldots,n$ are known and
fixed. Thus, the aim of the priority deriving procedure in the HRE
method is to calculate $w(a_{1}),\ldots,w(a_{k})$ based on the pairwise
comparisons matrix and the already known priorities of the reference
alternatives. 

As for AHP, we defined two methods for deriving priorities: the arithmetic
method \citep{Kulakowski2014hreaNoURL,Kulakowski2015ahre} and the
geometric method \citep{Kulakowski2015hreg}. To calculate the ranking
in the arithmetic HRE, we need to solve the equation: 
\[
Aw=b,
\]
where 
\[
A=\left[\begin{array}{cccc}
1 & -\frac{1}{n-1}c_{12} & \cdots & -\frac{1}{n-1}c_{1k}\\
-\frac{1}{n-1}c_{21} & 1 & \vdots & -\frac{1}{n-1}c_{2k}\\
\vdots & \vdots & \ddots & -\frac{1}{n-1}c_{k-1,k}\\
-\frac{1}{n-1}c_{k1} & \cdots & -\frac{1}{n-1}c_{k,k-1} & 1
\end{array}\right]
\]
is a $k\times k$ auxiliary matrix, and

\[
b=\left[\begin{array}{c}
\frac{1}{n-1}\sum^{n}_{j=k+1}c_{1j}w(a_{j})\\
\frac{1}{n-1}\sum^{n}_{j=k+1}c_{2j}w(a_{j})\\
\\\frac{1}{n-1}\sum^{n}_{j=k+1}c_{kj}w(a_{j})
\end{array}\right]
\]
is a constant term vector. In the case of a geometric approach, the
equation 
\[
\widehat{A}\widehat{w}=b
\]
is solved, where 

\[
\widehat{A}=\left[\begin{array}{cccc}
(n-1) & -1 & \cdots & -1\\
-1 & (n-1) & \cdots & -1\\
-1 & -1 & \ddots & -1\\
-1 & -1 & \cdots & (n-1)
\end{array}\right]
\]
 is a $k\times k$ matrix, 

\[
\widehat{A}\widehat{w}=\left[\begin{array}{c}
(n-1)\widehat{w}(a_{1})-\sum^{k}_{i=2}\widehat{w}(a_{i})\\
(n-1)\widehat{w}(a_{2})-\sum^{k}_{i=1,i\neq2}\widehat{w}(a_{i})\\
\vdots\\
(n-1)\widehat{w}(a_{k})-\sum^{k-1}_{i=1}\widehat{w}(a_{i})
\end{array}\right]
\]
 and
\[
b=\left[\begin{array}{c}
\sum^{n}_{i=2}lnc_{1j}+\sum^{n}_{i=k+1}\widehat{w}(a_{i})\\
\sum^{n}_{i=1,i\neq2}lnc_{2j}+\sum^{n}_{i=k+1}\widehat{w}(a_{i})\\
\vdots\\
\sum^{n}_{i=1,i\neq k}lnc_{kj}+\sum^{n}_{i=k+1}\widehat{w}(a_{i})
\end{array}\right].
\]

Since $\widehat{w}(a_{i})$ is the logarithmized value of $w(a_{i})$,
i.e., $\widehat{w}(a_{i})=\log_{e}w(a_{i})$, then the final weight
vector is obtained from $\widehat{w}$ by the exponential transformation
$w(a_{i})=e^{\widehat{w}(a_{i})}$ for $i=1,\ldots,k$. The solution
to a geometric (complete) HRE always exists and is optimal \citep{Kulakowski2015hreg}.
In the case of the arithmetic approach, a sufficiently small local
inconsistency is a sufficient condition \citep{Kulakowski2016note}.
\begin{example}
Consider the following situation. A company producing washing powders
is preparing a new advertising campaign. The company plans to print
billboards that will be displayed in prominent locations throughout
the country. Since this is an expensive operation, preliminary market
research is conducted. The company employs graphic designers who develop
three new advertising concepts. In addition, the company already has
two designs that were previously used in a smaller-scale campaign.
Market research data are available for these two existing designs
and show how many times the sales of washing powders increased after
each of them was displayed. The goal is therefore to rank all five
billboard designs and identify the best one. To this end, the company
forms a group of experts tasked with evaluating which graphics are
most eye-catching and memorable to consumers. The experts compare
the new designs pairwise and express their assessments numerically.
However, they do not need to compare the two existing designs, as
these serve as reference alternatives.

Let us denote $A_{U}=\{a_{1},a_{2},a_{3}\}$ as the set of new advertising
designs and $A_{K}=\{a_{4},a_{5}\}$ as the set of previously used
designs. According to the market research results, the \textquotedblleft profitability\textquotedblright{}
of designs $4$ and $5$ equals $w(a_{4})=6$ and $w(a_{5})=3$. The
pairwise assessments provided by the experts are recorded in the matrix
$C$: \begingroup
\renewcommand{\arraystretch}{1.4}

\[
C=\left[\begin{array}{ccccc}
1 & \frac{4}{3} & \frac{1}{3} & \frac{1}{9} & \frac{1}{6}\\
\frac{3}{4} & 1 & \frac{2}{5} & \frac{1}{8} & \frac{3}{7}\\
3 & \frac{5}{2} & 1 & 1 & \frac{8}{9}\\
9 & 8 & 1 & 1 & 2\\
6 & \frac{7}{3} & \frac{9}{8} & \frac{1}{2} & 1
\end{array}\right].
\]

\endgroup  To compute the weight function using the arithmetic HRE
method, we need to solve the following linear system 
\[
Aw=b,
\]
where
\[
A=\left[\begin{array}{ccc}
1 & -\frac{1}{3} & -\frac{1}{12}\\
-\frac{3}{16} & 1 & -\frac{1}{10}\\
-\frac{3}{4} & -\frac{5}{8} & 1
\end{array}\right],\,\,\,b=\left[\begin{array}{c}
0.291\\
0.508\\
2.166
\end{array}\right],
\]
thus 
\[
\left[\begin{array}{c}
w(a_{1})\\
w(a_{2})\\
w(a_{3})
\end{array}\right]=\left[\begin{array}{c}
0.928\\
1.034\\
3.509
\end{array}\right].
\]
To compute the ranking function using the geometric HRE method, we
solve the system $\widehat{A}\,\widehat{w}=b$, where 
\[
\widehat{A}=\left[\begin{array}{ccc}
4 & -1 & -1\\
-1 & 4 & -1\\
-1 & -1 & 4
\end{array}\right],\,\,\,b=\left[\begin{array}{c}
-0.829\\
-0.538\\
2.079
\end{array}\right].
\]
The resulting weight vectors are 
\[
\left[\begin{array}{c}
\widehat{w}(a_{1})\\
\widehat{w}(a_{2})\\
\widehat{w}(a_{3})
\end{array}\right]=\left[\begin{array}{c}
-0.218\\
-0.084\\
1.121
\end{array}\right],\,\,\,\left[\begin{array}{c}
w(a_{1})\\
w(a_{2})\\
w(a_{3})
\end{array}\right]=\left[\begin{array}{c}
0.804\\
0.919\\
3.068
\end{array}\right].
\]
\end{example}

\section{Arithmetic Incomplete Heuristic Rating Estimation\label{sec:Additive-Incomplete-Heuristic}}

As in the original method, we assume that the set of alternatives
$A=\{a_{1},\ldots,a_{n}\}$ consists of two subsets $A_{U}=\{a_{1},\ldots,a_{k}\}$
and $A_{K}=\{a_{k+1},\ldots,a_{n}\}$. $A_{K}$ contains references
(i.e., alternatives with the known priority values), and $A_{U}$
is composed of alternatives whose ranking has yet to be calculated. 

Since it is natural to expect that $w(a_{i})$ is similar\footnote{Ideally $w(a_{i})=c_{ij}w(a_{j})$. This is when the PC matrix $C$
is consistent.} to $c_{ij}w(a_{j})$, i.e.

\begin{equation}
w(a_{i})\approx c_{ij}w(a_{j})\label{eq:fundamental-assumption}
\end{equation}
then, we may approximate $w(a_{i})$ by the weighted average of all
other priority values: 
\begin{equation}
w(a_{i})=\frac{1}{n-1}\sum^{n}_{\substack{j=1\\
i\neq j
}
}c_{ij}w(a_{j}),\,\,\text{for}\,\,i=1,\ldots,n.\label{eq:average-postulate}
\end{equation}
However, when $c_{ij}$ is undefined i.e. $c_{ij}=?$ we assume, following
the implementation in the incomplete EVM and GMM/LLSM methods \citep{Harker1987amoq,kulakowski2020otgm,Bozoki2010ooco},
that 
\[
c_{ij}=\frac{w(a_{i})}{w(a_{j})}.
\]
In other words, instead of an incomplete matrix 
\begin{equation}
C=\left[\begin{array}{ccccc}
1 & c_{12} & c_{13} & \cdots & c_{1n}\\
c_{21} & 1 & c_{23} & \cdots & c_{2n}\\
\vdots &  & \vdots & \vdots & \vdots\\
\vdots & \vdots & \vdots & \vdots & \vdots\\
c_{n1} & c_{n2} & c_{n3} & \cdots & 1
\end{array}\right],\label{eq:incomplete-matrix-in-hre}
\end{equation}
we will use the supplemented matrix 

\begin{equation}
C^{*}=\left[\begin{array}{ccccc}
1 & g_{12} & g_{13} & \cdots & g_{1n}\\
g_{21} & 1 & g_{23} & \cdots & g_{2n}\\
\vdots &  & \vdots & \vdots & \vdots\\
\vdots & \vdots & \vdots & \vdots & \vdots\\
g_{n1} & g_{n2} & g_{n3} & \cdots & 1
\end{array}\right],\label{eq:complemented-matrix}
\end{equation}
where 
\[
g_{ij}=\begin{cases}
c_{ij} & \text{if}\,\,c_{ij}\neq?\\
\frac{w(a_{i})}{w(a_{j})} & \text{if}\,\,c_{ij}=?
\end{cases}.
\]

In other words, we assume that the missing values are consistent with
the final ranking result, which is yet to be determined. As a result,
(\ref{eq:average-postulate}) takes the form: 

\[
w(a_{i})=\frac{1}{n-1}\left(\sum_{\substack{j=1,i\neq j\\
c_{ij}\neq?
}
}c_{ij}w(a_{j})+\sum_{\substack{j=1,i\neq j\\
c_{ij}=?
}
}w(a_{i})\right),\,\,\text{for}\,\,i=1,\ldots,n.
\]
Thus, let $s_{i}$ be the number of undefined comparisons between
$a_{i}$ and $a_{j}\in A_{U}$ and $r_{i}$ be the number of undefined
comparisons between $a_{i}$ and $a_{j}\in A_{K}$, hence the above
takes the form 

\[
w(a_{i})=\frac{1}{n-1}\left(\sum_{\substack{j=1,i\neq j\\
c_{ij}\neq?
}
}c_{ij}w(a_{j})+(s_{i}+r_{i})w(a_{i})\right),\,\,\text{for}\,\,i=1,\ldots,n.
\]

By multiplying both sides by $n-1$ we get 

\[
(n-1)w(a_{i})=\sum_{\substack{j=1,i\neq j\\
c_{ij}\neq?
}
}c_{ij}w(a_{j})+(s_{i}+r_{i})w(a_{i}),\,\,\text{for}\,\,i=1,\ldots,n,
\]

i.e. 
\[
(n-s_{i}-r_{i}-1)w(a_{i})=\sum_{\substack{j=1,i\neq j\\
c_{ij}\neq?
}
}c_{ij}w(a_{j}),\,\,\text{for}\,\,i=1,\ldots,n,
\]

and finally 
\[
w(a_{i})=\frac{1}{n-s_{i}-r_{i}-1}\sum_{\substack{j=1,i\neq j\\
c_{ij}\neq?
}
}c_{ij}w(a_{j}),\,\,\text{for}\,\,i=1,\ldots,n.
\]

The above can be written in the form of an equation system:

\[
\begin{array}{ccc}
w(a_{1}) & = & \frac{1}{n-s_{1}-r_{i}-1}(d_{1,2}w(a_{2})+\dotfill+d_{1,n}w(a_{n}))\\
w(a_{2}) & = & \frac{1}{n-s_{2}-r_{i}-1}(d_{2,1}w(a_{1})+d_{2,3}w(a_{3})+\dotfill+d_{2,n}w(a_{n}))\\
\hdotsfor[1]{3}\\
w(a_{k}) & = & \frac{1}{n-s_{k}-r_{i}-1}\left(d_{k,1}w(a_{1})+\ldots+d_{k,k-1}w(a_{k-1})+d_{k,k+1}w(a_{k+1})+\right.\\
 &  & \left.+d_{k,n}w(a_{n})\right)
\end{array},
\]

where 
\[
d_{ij}=\begin{cases}
c_{ij} & \text{if}\,\,c_{ij}\neq?\\
0 & \text{if}\,\,c_{ij}=?
\end{cases}.
\]

The values $w(a_{k+1}),\ldots,w(a_{n})$ are known and constant ($a_{k+1},\ldots,a_{n}$
belongs to $A_{K}$, the set of known alternatives). For the same
reason, all the comparisons $c_{ij}$ where $i,j=k+1,\ldots n$ do
not need to be determined by experts. Their values are: 
\[
c_{k+1,k+2}=\frac{w(a_{k+1})}{w(a_{k+2})},\ldots,c_{n,n-1}=\frac{w(a_{n})}{w(a_{n-1})}.
\]
Therefore, we can sum all the components in the row composed of known
elements. Let us denote:

\[
b_{j}=\frac{1}{n-s_{j}-r_{i}-1}c_{j,k+1}w(a_{k+1})+\ldots+\frac{1}{n-s_{j}-r_{i}-1}c_{j,n}w(a_{n}).
\]
Thus, we can write the linear equations system as:

\[
\begin{array}{ccc}
w(a_{1}) & = & \frac{1}{n-s_{1}-r_{1}-1}d_{1,2}w(a_{2})+\dotfill+\frac{1}{n-s_{1}-r_{1}-1}d_{k,1}w(a_{k})+b_{1}\\
w(a_{2}) & = & \frac{1}{n-s_{2}-r_{2}-1}d_{2,1}w(a_{1})+\dotfill+\frac{1}{n-s_{2}-r_{2}-1}d_{2,k}w(a_{k})+b_{2}\\
\hdotsfor[1]{3}\\
w(a_{k}) & = & \frac{1}{n-s_{k}-r_{k}-1}d_{k,1}w(a_{1})+\dotfill+\frac{1}{n-s_{k}-r_{k}-1}d_{k,k-1}w(a_{k-1})+b_{k}
\end{array}.
\]
The matrix form of the above equation system is given as
\begin{equation}
\overline{C}w=b,\label{eq:hre-addit-eq-system}
\end{equation}
where \begingroup
\setlength{\arraycolsep}{-7pt}
\renewcommand{\arraystretch}{1.4}

\begin{equation}
\overline{C}=\left(\begin{array}{cccc}
1 & -\frac{1}{n-s_{1}-r_{1}-1}d_{1,2} & \cdots & -\frac{1}{n-s_{1}-r_{1}-1}d_{1,k}\\
-\frac{1}{n-s_{2}-r_{2}-1}d_{2,1} & 1 & \cdots & -\frac{1}{n-s_{2}-r_{2}-1}d_{2,k}\\
\vdots & \vdots & \vdots & \vdots\\
\,\,-\frac{1}{n-s_{k-1}-r_{k-1}-1}d_{k-1,1} & \cdots & \ddots & -\frac{1}{n-s_{k-1}-r_{k-1}-1}d_{k-1,k}\,\,\\
-\frac{1}{n-s_{k}-r_{k}-1}d_{k,1} & \cdots & -\frac{1}{n-s_{k}-r_{k}-1}d_{k,k-1} & 1
\end{array}\right),\label{eq:25-eq-1}
\end{equation}

\endgroup 

the vector of constant terms is \begingroup
\renewcommand{\arraystretch}{1.4} 
\begin{equation}
b=\left(\begin{array}{c}
\frac{1}{n-s_{1}-r_{1}-1}c_{1,k+1}w(a_{k+1})+\ldots+\frac{1}{n-s_{1}-r_{1}-1}c_{1,n}w(a_{n})\\
\frac{1}{n-s_{2}-r_{2}-1}c_{2,k+1}w(a_{k+1})+\ldots+\frac{1}{n-s_{2}-r_{2}-1}c_{2,n}w(a_{n})\\
\vdots\\
\frac{1}{n-s_{k}-r_{k}-1}c_{k,k+1}w(a_{k+1})+\ldots+\frac{1}{n-s_{k}-r_{k}-1}c_{k,n}w(a_{n})
\end{array}\right),\label{eq:constant_terms_vector}
\end{equation}
\endgroup  and values that need to be determined are denoted as:

\begin{equation}
w=\left(\begin{array}{c}
w(a_{1})\\
\vdots\\
\vdots\\
w(a_{k})
\end{array}\right).\label{eq:unknown_terms_vector}
\end{equation}

\begin{example}
\label{subsec:Numerical-example-1}We are planning a formal party
with refreshments. We have already selected the main course and dessert,
and now we would like to choose a tasty appetizer. In addition to
two well-tested options \textemdash{} fish soup and a cheese and charcuterie
board \textemdash{} we are considering four other appetizers of similar
price: portobello mushrooms baked with halloumi, a paprika-stuffed
tortilla, avocado with salmon, and a gyros salad with pasta. Before
the party, our guests act as testers. Since we only serve two dishes
to them at a time, it is not possible to directly compare all the
appetizers. Based on the pairwise evaluations provided by our guests,
we construct a hierarchy of the appetizers.

Let $A_{U}=\{a_{1}-\text{mushrooms},\,a_{2}-\text{tortilla},\,a_{3}-\text{avocado},\,a_{4}-\text{gyros}\}$
and $A_{K}=\{a_{5}-\text{soup},a_{6}-\text{cheese}\}$. A PC-matrix
containing appetizer comparisons is presented below \begingroup
\renewcommand{\arraystretch}{1.4} 

\begin{equation}
C=\left[\begin{array}{cccccc}
1 & \frac{3}{2} & ? & 3 & 2 & 3\\
\frac{2}{3} & 1 & 2 & ? & ? & 1\\
? & \frac{1}{2} & 1 & \frac{1}{2} & \frac{1}{3} & \frac{5}{3}\\
\frac{1}{3} & ? & 2 & 1 & \frac{2}{3} & ?\\
\frac{1}{2} & ? & 3 & \frac{3}{2} & 1 & \frac{3}{2}\\
\frac{1}{3} & 1 & \frac{3}{5} & ? & \frac{2}{3} & 1
\end{array}\right].\label{eq:example_pc_incomplete_matrix}
\end{equation}

\endgroup  If two dishes are not compared, we will mark it with a
$?$ sign in the matrix $C$. In further calculations, the characters
$?$ are replaced with the expression $c_{ij}=\frac{w(a_{i})}{w(a_{j})}$,
thus \begingroup
\renewcommand{\arraystretch}{1.4} 
\[
C=\left[\begin{array}{cccccc}
1 & \frac{3}{2} & \frac{w(a_{1})}{w(a_{3})} & 3 & 2 & 3\\
\frac{2}{3} & 1 & 2 & \frac{w(a_{2})}{w(a_{4})} & \frac{w(a_{2})}{w(a_{5})} & 1\\
\frac{w(a_{3})}{w(a_{1})} & \frac{1}{2} & 1 & \frac{1}{2} & \frac{1}{3} & \frac{5}{3}\\
\frac{1}{3} & \frac{w(a_{4})}{w(a_{2})} & 2 & 1 & \frac{2}{3} & \frac{w(a_{4})}{w(a_{6})}\\
\frac{1}{2} & \frac{w(a_{5})}{w(a_{2})} & 3 & \frac{3}{2} & 1 & \frac{3}{2}\\
\frac{1}{3} & 1 & \frac{3}{5} & \frac{w(a_{6})}{w(a_{4})} & \frac{2}{3} & 1
\end{array}\right].
\]
 \endgroup  The value $w(a_{5})$ and $w(a_{6})$ are known and set
to $6$ and $4$ points correspondingly. So it corresponds to the
following equation system:
\[
\begin{array}{c}
w(a_{1})=\frac{1}{5}\left(\frac{3}{2}w(a_{2})+w(a_{1})+3w(a_{4})+2w(a_{5})+3w(a_{6})\right)\\
w(a_{2})=\frac{1}{5}\left(\frac{2}{3}w(a_{1})+2w(a_{3})+w(a_{2})+w(a_{2})+w(a_{6})\right)\\
w(a_{3})=\frac{1}{5}\left(w(a_{3})+\frac{1}{2}w(a_{2})+\frac{1}{2}w(a_{4})+\frac{1}{3}w(a_{5})+\frac{5}{3}w(a_{6})\right)\\
w(a_{4})=\frac{1}{5}\left(\frac{1}{3}w(a_{1})+w(a_{4})+2w(a_{3})+\frac{2}{3}w(a_{5})+w(a_{4})\right)
\end{array}.
\]
Multiplying both sides of each equation by $5$ and appropriately
grouping we get a 
\[
\begin{array}{c}
4w(a_{1})-\frac{3}{2}w(a_{2})-0w(a_{3})-3w(a_{4})=2w(a_{5})+3w(a_{6})\\
-\frac{2}{3}w(a_{1})+3w(a_{2})-2w(a_{3})-0w(a_{4})=w(a_{6})\\
-0w(a_{1})-\frac{1}{2}w(a_{2})+4w(a_{3})-\frac{1}{2}w(a_{4})=\frac{1}{3}w(a_{5})+\frac{5}{3}w(a_{6})\\
-\frac{1}{3}w(a_{1})-0w(a_{2})-2w(a_{3})+3w(a_{4})=\frac{2}{3}w(a_{5})
\end{array},
\]

which is equivalent to
\[
\begin{array}{c}
w(a_{1})-\frac{3}{8}w(a_{2})-\frac{0}{4}w(a_{3})-\frac{3}{4}w(a_{4})=\frac{1}{2}w(a_{5})+\frac{3}{4}w(a_{6})=6\\
-\frac{2}{9}w(a_{1})+w(a_{2})-\frac{2}{3}w(a_{3})-\frac{0}{3}w(a_{4})=\frac{1}{3}w(a_{6})=\frac{4}{3}\\
-\frac{0}{4}w(a_{1})-\frac{1}{8}w(a_{2})+w(a_{3})-\frac{1}{8}w(a_{4})=\frac{1}{12}w(a_{5})+\frac{5}{12}w(a_{6})=\frac{13}{6}\\
-\frac{1}{9}w(a_{1})-0w(a_{2})-\frac{2}{3}w(a_{3})+w(a_{4})=\frac{2}{9}w(a_{5})=\frac{4}{3}
\end{array}.
\]

So eventually, the system of equations takes the form
\[
\overline{C}w=b,
\]
where \begingroup
\setlength{\arraycolsep}{3pt}
\renewcommand{\arraystretch}{1.4}
\[
\overline{C}=\left[\begin{array}{cccc}
1 & -\frac{3}{8} & 0 & -\frac{3}{4}\\
-\frac{2}{9} & 1 & -\frac{2}{3} & 0\\
0 & -\frac{1}{8} & 1 & -\frac{1}{8}\\
-\frac{1}{9} & 0 & -\frac{2}{3} & 1
\end{array}\right],
\]
and
\[
b=\left[\begin{array}{c}
6\\
\frac{4}{3}\\
\frac{13}{6}\\
\frac{4}{3}
\end{array}\right].
\]

\endgroup

Solving the above equation, we obtain the weight vector $\overline{C}^{-1}b=w$,
where $w=\left[12.251,6.464,3.612,5.102\right]^{T}$. After adding
reference values $w(a_{5})=6$ and $w(a_{6})=4$, 
\[
w_{\text{aihre}}=\left[12.251,6.464,3.612,5.102,6,4\right]^{T},
\]
and finally, after normalization, we obtain a rescaled weight vector:
\[
w_{\text{aihre}}/\left\Vert w_{\text{aihre}}\right\Vert _{1}=\left[0.327,\,0.172,\,0.0965,\,0.136,\,0.16,\,0.107\right]^{T}.
\]
The options, from best to worst, are: $a_{1}$, $a_{2}$, $a_{5}$,
$a_{4}$, $a_{6}$ and $a_{3}$. Based on the ranking results, we
selected the top three options: $a_{1}-\text{mushrooms}$, $a_{2}-\text{tortilla}$
and $a_{5}-\text{soup}.$
\end{example}

\section{Geometric Incomplete Heuristic Rating Estimation\label{sec:Multiplicative-Incomplete-Heuris}}

With the same assumptions about the sets $A,A_{U}$ and $A_{K}$,
we may want to request that  $w(a_{i})$ be the geometric mean of
the components $c_{ij}w(a_{j})$ for $j=1,\ldots,n$. This leads to
a nonlinear system of equations: 

\begin{equation}
w(a_{i})=\left(\prod^{n}_{\substack{j=1\\
i\neq j
}
}c_{ij}w(a_{j})\right)^{\frac{1}{n-1}},\,\,\text{for}\,\,i=1,\ldots,n\label{eq:geom-nonlinear-eq-system}
\end{equation}
Once again, having in mind (\ref{eq:fundamental-assumption}), when
$c_{ij}$ is undefined, i.e., $c_{ij}=?$ we assume that 
\[
c_{ij}=\frac{w(a_{i})}{w(a_{j})}.
\]
Thus, the above (\ref{eq:geom-nonlinear-eq-system}) takes the form

\begin{equation}
w(a_{i})=\left(w^{s_{i}+r_{i}}(a_{i})\prod^{n}_{\substack{j=1,i\neq j\\
c_{ij}\neq?
}
}c_{ij}w(a_{j})\right)^{\frac{1}{n-1}},\,\,\text{for}\,\,i=1,\ldots,n,\label{eq:incompl-geometric-hre-eq}
\end{equation}
where $s_{i}$ is the number of missing comparisons for the i-th alternative
with alternatives from $A_{U}$, and $r_{i}$ is the number of missing
comparisons for the i-th alternative with alternatives from $A_{K}$.
After raising both sides to the power of $n-1$, we get a nonlinear
equation system: 

\begin{equation}
\begin{array}{ccc}
w^{n-1}(a_{1}) & = & 1\cdot d_{1,2}\cdot\dotfill\cdot d_{1,n}\cdot w^{s_{1}+r_{1}}(a_{1})\\
w^{n-1}(a_{2}) & = & d_{2,1}\cdot1\cdot d_{2,3}\cdot\dotfill\cdot d_{2,n}\cdot w^{s_{2}+r_{2}}(a_{2})\\
\hdotsfor[1]{3}\\
w^{n-1}(a_{k}) & = & d_{k,1}\cdot\dotfill\cdot d_{k,n-1}\cdot1\cdot w^{s_{k}+r_{k}}(a_{k})
\end{array},\label{eq:eq-system-1}
\end{equation}
where 
\[
d_{ij}=\begin{cases}
c_{ij}w(a_{j}) & \text{if}\,\,c_{ij}\neq?\\
1 & \text{if}\,\,c_{ij}=?
\end{cases}.
\]
Thus the equation system (\ref{eq:eq-system-1}) can be written as

\begin{equation}
\begin{array}{ccc}
w^{n-s_{1}-r_{1}-1}(a_{1}) & = & 1\cdot d_{1,2}\cdot\dotfill\cdot d_{1,n}\\
w^{n-s_{2}-r_{2}-1}(a_{2}) & = & d_{2,1}\cdot1\cdot d_{2,3}\cdot\dotfill\cdot d_{2,n}\\
\hdotsfor[1]{3}\\
w^{n-s_{k}-r_{k}-1}(a_{k}) & = & d_{k,1}\cdot\dotfill\cdot d_{k,n-1}\cdot1
\end{array}.\label{eq:eq-system-2}
\end{equation}
As the values $w(a_{k+1}),\ldots,w(a_{n})$ are fixed and known, and
as follows, $c_{ij}w(a_{j})$ are initially known and constant, let
us denote: 
\[
g_{i}=\prod^{n}_{\stackrel{j=k+1}{c_{ij}\neq?}}c_{ij}w(a_{j}).
\]
Hence, the equation system (\ref{eq:eq-system-2}) can be written
as: 

\begin{equation}
\begin{array}{ccc}
w^{n-s_{1}-r_{1}-1}(a_{1}) & = & 1\cdot d_{1,2}\cdot\dotfill\cdot d_{1,k}\cdot g_{1}\\
w^{n-s_{2}-r_{2}-1}(a_{2}) & = & d_{2,1}\cdot1\cdot d_{2,3}\cdot\dotfill\cdot d_{2,k}\cdot g_{2}\\
\hdotsfor[1]{3}\\
w^{n-s_{k}-r_{k}-1}(a_{k}) & = & d_{k,1}\cdot\dotfill\cdot d_{k,k-1}\cdot g_{k}
\end{array}.\label{eq:eq-system-3}
\end{equation}
By logarithmic transformation of (\ref{eq:eq-system-3}), we get 

\[
\begin{array}{ccc}
(n-s_{1}-r_{1}-1)\widehat{w}(a_{1}) & = & \widehat{d}_{1,2}+\dotfill+\widehat{d}_{1,k}+\widehat{g}_{1}\\
(n-s_{2}-r_{2}-1)\widehat{w}(a_{2}) & = & \widehat{d}_{2,1}+\dotfill+\widehat{d}_{2,k}+\widehat{g}_{2}\\
\hdotsfor[1]{3}\\
(n-s_{k}-r_{k}-1)\widehat{w}(a_{k}) & = & \widehat{d}_{k,1}+\ldots+\widehat{d}_{k,k-1}+\widehat{g}_{k}
\end{array}.
\]
where $\log_{\xi}w(a_{i})\overset{\textit{df}}{=}\widehat{w}(a_{i})$,
$\log_{\xi}c_{ij}\overset{\textit{df}}{=}\widehat{c}_{ij}$, $\log_{\xi}d_{ij}\overset{\textit{df}}{=}\widehat{d}_{ij}$
and $\log_{\xi}g_{j}\overset{\textit{df}}{=}\widehat{g}_{j}$ for
some real constant $\xi\in\mathbb{R}_{+}$. Note that if $c_{ij}=?$
then corresponding $\widehat{d}_{ij}=0$. When $c_{ij}\neq?$ then
$\widehat{d}_{ij}=\widehat{w}(a_{j})+\widehat{c}_{ij}$. 

Let us group the constant values in each row. As a result, we obtain:

\[
\begin{array}{ccc}
(n-s_{1}-r_{1}-1)\widehat{w}(a_{1})-\sum^{k}_{j=2,c_{1,j}\neq?}\widehat{w}(a_{j})\,\,\,\,\,\,\,\,\,\,\, & = & b_{1}\\
(n-s_{2}-r_{2}-1)\widehat{w}(a_{2})-\sum^{k}_{j=1,j\neq2,c_{2j}\neq?}\widehat{w}(a_{j}) & = & b_{2}\\
\hdotsfor[1]{3}\\
(n-s_{k}-r_{k}-1)\widehat{w}(a_{k})-\sum^{k-1}_{j=1,c_{kj}\neq?}\widehat{w}(a_{j})\,\,\,\,\,\,\,\,\,\, & = & b_{k}
\end{array},
\]
where $b_{i}\overset{\textit{df}}{=}\sum^{k}_{j=1,j\neq i,c_{ij}\neq?}\widehat{c}_{i,j}+\widehat{g}_{i}$
for $i=1,\ldots,k$. 

We can write the above linear equation system in the form of the following
matrix equation:
\begin{equation}
\widehat{C}\widehat{w}=b,\label{eq:hre-geom-matrix-incompl}
\end{equation}

where 
\[
\widehat{C}=\left[\begin{array}{cccc}
(n-s_{1}-r_{1}-1) & q_{1,2} & \cdots & q_{1,k}\\
\vdots & \ddots &  & \vdots\\
\vdots &  & \ddots & \vdots\\
q_{k,1} & q_{k,2} & \cdots & (n-s_{k}-r_{k}-1)
\end{array}\right],
\]

with 
\begin{equation}
q_{ij}=\begin{cases}
-1 & \text{if}\,\,c_{ij}\neq?\\
0 & \text{if}\,\,c_{ij}=?
\end{cases},\label{eq:q_i}
\end{equation}

and 
\[
\widehat{w}=\left[\begin{array}{c}
\widehat{w}(a_{1})\\
\widehat{w}(a_{2})\\
\vdots\\
\widehat{w}(a_{k})
\end{array}\right],\,\,\,\text{and}\,\,\,b=\left[\begin{array}{c}
b_{1}\\
b_{2}\\
\vdots\\
b_{k}
\end{array}\right].
\]

By solving (\ref{eq:hre-geom-matrix-incompl}), we get the solution
of the original nonlinear equation (\ref{eq:eq-system-1}). Indeed,
when computed, the original priority vector is given as: 
\begin{equation}
w=\left(\begin{array}{c}
\xi^{\widehat{w}(a_{1})}\\
\vdots\\
\vdots\\
\xi^{\widehat{w}(a_{k})}
\end{array}\right).\label{eq:final solution geom}
\end{equation}

\begin{example}
For comparison, let us consider the same example as before. In our
model, we therefore have four non-referential alternatives, i.e.,
$A_{U}=\{a_{1},\ldots,a_{4}\}$ and two reference alternatives $A_{K}=\{a_{5},a_{6}\}$.
The weight values for $a_{5}$ and $a_{6}$ are $6$ and $4$, respectively.
The pairwise comparison matrix remains unchanged (see \ref{eq:example_pc_incomplete_matrix}).
Following the procedure outlined above, we create the following system
of equations:
\[
\begin{array}{c}
w^{5}(a_{1})=\frac{3}{2}w(a_{2})\cdot w(a_{1})\cdot3w(a_{4})\cdot2w(a_{5})\cdot3w(a_{6})\\
w^{5}(a_{2})=\frac{2}{3}w(a_{1})\cdot2w(a_{3})\cdot w(a_{2})\cdot w(a_{2})\cdot w(a_{6})\\
w^{5}(a_{3})=w(a_{3})\cdot\frac{1}{2}w(a_{2})\cdot\frac{1}{2}w(a_{4})\cdot\frac{1}{3}w(a_{5})\cdot\frac{5}{3}w(a_{6})\\
w^{5}(a_{4})=\frac{1}{3}w(a_{1})\cdot w(a_{4})\cdot2w(a_{3})\cdot\frac{2}{3}w(a_{5})\cdot w(a_{4})
\end{array}
\]

and then, after reducing the terms on both sides of each equation
and substituting $\ln w(a_{5})=\widehat{w}(a_{5})$ (we assume $\xi=e$),
we take the logarithm of both sides, 
\[
\begin{array}{c}
4\widehat{w}(a_{1})=\ln\frac{3}{2}+\widehat{w}(a_{2})+\widehat{w}(a_{4})+\ln2+\widehat{w}(a_{5})+2\ln3+\widehat{w}(a_{6})\\
3\widehat{w}(a_{2})=\ln\frac{2}{3}+\widehat{w}(a_{1})+\ln2+\widehat{w}(a_{3})+\widehat{w}(a_{6})\\
4\widehat{w}(a_{3})=\ln\frac{1}{2}+\widehat{w}(a_{2})+\ln\frac{1}{2}+\widehat{w}(a_{4})+\ln\frac{1}{3}+\widehat{w}(a_{5})+\ln\frac{5}{3}+\widehat{w}(a_{6})\\
3\widehat{w}(a_{4})=\ln\frac{1}{3}+\widehat{w}(a_{1})+\ln2+\widehat{w}(a_{3})+\ln\frac{2}{3}+\widehat{w}(a_{5})
\end{array}
\]

After merging similar words and regrouping, we obtain

\[
\begin{array}{c}
4\widehat{w}(a_{1})-\widehat{w}(a_{2})-0\widehat{w}(a_{3})-\widehat{w}(a_{4})=3\ln3+\widehat{w}(a_{5})+\widehat{w}(a_{6})=\ln648\\
-\widehat{w}(a_{1})+3\widehat{w}(a_{2})-\widehat{w}(a_{3})-0\widehat{w}(a_{4})=+2\ln2-\ln3+\widehat{w}(a_{6})=\ln\frac{16}{3}\\
0\widehat{w}(a_{1})-\widehat{w}(a_{2})+4\widehat{w}(a_{3})-w(a_{4})=\widehat{w}(a_{5})-\ln\frac{36}{5}+\widehat{w}(a_{6})=\ln\frac{10}{3}\\
-\widehat{w}(a_{1})+0\widehat{w}(a_{2})-\widehat{w}(a_{3})+3\widehat{w}(a_{4})=\ln\frac{1}{3}+\ln2+\ln\frac{2}{3}+\widehat{w}(a_{5})=\ln\frac{8}{3}
\end{array}.
\]

This leads to the equation $\widehat{C}\widehat{w}=b$ where \begingroup
\renewcommand{\arraystretch}{1.4} 
\[
\widehat{C}=\left[\begin{array}{cccc}
4 & -1 & 0 & -1\\
-1 & 3 & -1 & 0\\
0 & -1 & 4 & -1\\
-1 & 0 & -1 & 3
\end{array}\right],\,\,\text{and}\,\,b=\left[\begin{array}{c}
\ln648\\
\ln\frac{16}{3}\\
\ln\frac{10}{3}\\
\ln\frac{8}{3}
\end{array}\right].
\]

\endgroup  An intermediate solution $\widehat{w}=\widehat{C}^{-1}b$
is as follows 
\[
\widehat{w}=\left[\{2.43,1.738,1.112,1.507\right]^{T}.
\]

Therefore, after raising constant $e$ to the powers of the vector
elements $\widehat{w}$ and supplementing with reference values, we
obtain

\[
w_{\text{mhre}}=\left[11.49,5.82,3.22,4.621,6,4\right]^{T},
\]

and finally, after normalization

\[
w_{\text{mhre}}/\left\Vert w_{\text{mhre}}\right\Vert _{1}=\left[0.326,0.165,0.091,0.131,0.17,0.113\right]^{T}.
\]

Although the calculated priorities of individual alternatives are
slightly different from those in Example \ref{subsec:Numerical-example-1},
their order remains unchanged. As in the previous case, we choose
the first three options, $a_{1}$, $a_{2}$, and $a_{5}$ as snacks
for the planned party.
\end{example}

\section{Solution of Incomplete Heuristic Rating Estimation\label{sec:Solution-of-Incomplete}}

Both incomplete HRE methods, the arithmetic and geometric approaches,
are based on solving specific systems of linear equations. This raises
the question of under what conditions such systems of equations have
a solution. Below, we will show that in the arithmetic case, attempting
to find a weight vector by solving a system of equations is not always
successful. We will also prove that the geometric method does not
have these limitations.  

\subsection{Solution of arithmetic approach}

The matrix of the system of equations (\ref{eq:hre-addit-eq-system})
for the aiHRE problem has a rather specific form. Indeed, the matrix
$\bar{A}=[\bar{a}_{ij}]$ (\ref{eq:25-eq-1}) has a real positive
diagonal, while all elements outside the diagonal are real non-positive,
i.e., $\bar{a}_{ij}\leq0$ for $i\neq j$. This, combined with knowledge
of the problem's structure, allows us to formulate a sufficient condition
for the existence of an exact solution to the aiHRE problem. 
\begin{thm}
\label{prop:A-multiplicative-incomplete} An arithmetic incomplete
Heuristic Rating Estimation (aiHRE) problem for alternatives $A=A_{U}\cup A_{K}$
with equation $\bar{A}w=b$ (\ref{eq:hre-addit-eq-system}) has a
unique, real and positive solution (\ref{eq:final solution geom})
if $b\geq0$ and where there exists at least one $b_{i}$ such that
$b_{i}>0$, $\overline{C}$ is irreducible and 
\begin{equation}
\text{\ensuremath{\left|A_{K}\right|}}-r_{i}+z_{i}\geq\sum^{k}_{\stackrel{j=1}{c_{ij}\neq?}}c_{ij},\,\,\,\text{for}\,\,\,i=1,\ldots,k\label{eq:propozycja-istnienia-addytywna}
\end{equation}
where $z_{i}$ is the number of existing ($\neq?$) pairwise comparisons
in the i-th row of the matrix $\overline{C}$, and there is one such
row $q$ of the matrix $\overline{C}$ for which the inequality (\ref{eq:propozycja-istnienia-addytywna})
is strict.
\end{thm}
\begin{proof}
Considering the diagonally dominant matrix (Def. \ref{def:diag-dom-matrix})
$\bar{A}$ we get 
\[
\left|1\right|\ge\sum^{k}_{\stackrel{j=1}{i\neq j}}\left|\frac{1}{n-s_{i}-r_{i}-1}d_{ij}\right|\,\,\,\text{for}\,\,\,1\leq i\leq k.
\]

By dropping absolute values and the minus sign
\[
1\ge\sum^{k}_{j=1}\frac{1}{n-s_{i}-r_{i}-1}d_{ij},
\]

and denoting the number of comparisons in the i-th row by $z_{i}=k-s_{i}-1$
we receive
\[
n-s_{i}-r_{i}-1\geq\sum^{k}_{j=1}d_{ij}=\underset{z_{i}=k-s_{i}-1}{\underbrace{c_{ij_{1}}+c_{ij_{2}}\ldots+c_{ij_{k-s_{i}-1}}}}
\]

and finally as $\left|A_{K}\right|=n-k$
\begin{equation}
\left|A_{K}\right|+z_{i}-r_{i}\geq\sum^{k}_{\stackrel{j=1}{c_{ij}\neq?}}c_{ij}=z_{i}\,\,\,\text{for}\,\,\,1\leq i\leq k.\label{eq:final-proof-ineq}
\end{equation}

Therefore, if the above inequality is strict for at least one of the
rows, given that the above transformations were identity transformations,
due to the Corollaries \ref{cor:postivie-matrix-observ} and \ref{cor:m-matrix-corollary}
we obtain that $\overline{C}$ is a nonsingular M-matrix where $\overline{C}^{-1}>0$.
Since the constant term vector $b\geq0$, then the weight vector $w=\overline{C}^{-1}b$
is admissible, i.e., real and positive. 
\end{proof}

In the above statement, the assumptions concerning the irreducibility
of $\overline{C}$ and $b\geq0$ seem to be quite natural. The first
of these is the requirement that any two non-referential alternatives
be directly or indirectly (through a sequence of comparisons) comparable
to each other. A similar assumption regarding the irreducibility of
matrices also appears in the EVM and GMM methods for incomplete matrices
\citep{Harker1987amoq,kulakowski2020otgm}. The second common-sense
assumption, that $b\geq0$ (where $\exists b_{i}>0$), means in practice
that at least one non-reference alternative is directly compared with
at least one reference alternative. 

From the condition, it is easy to see that a large number of reference
alternatives favors the existence of a solution $\left|A_{K}\right|$
(the larger the better) and a small number of missing comparisons
with reference alternatives $r_{i}$. Limiting the scale of comparisons
also has a positive effect. Here, we should note that in practice,
for $C=[c_{ij}]$, its entries come from a certain range $c_{ij}\in[1/t,\,t]$.
That is, every $c_{ij}\leq t$. Therefore, from the proof of Proposition
\ref{prop:A-multiplicative-incomplete}, we have 
\[
\left|A_{K}\right|+z-r_{i}\geq t\cdot z_{i}\geq\sum^{k}_{\stackrel{j=1}{c_{ij}\neq?}}c_{ij}\,\,\,\text{for}\,\,\,1\leq i\leq k.
\]
Therefore, alternatively, we can demand that 
\[
\frac{\left|A_{K}\right|+z_{i}-r_{i}}{z_{i}}\geq t\,\,\,\,\text{for}\,\,\,1\leq i\leq k,
\]
whereby for one selected $i$, the above inequality should be strict.
The above formula clearly shows the relationship between the highest
degree of the scale (the maximum comparison value) and the number
of reference alternatives. In particular, it is clear that with a
fixed $t$, by increasing the number of reference alternatives (and
presumably the number of all alternatives in the decision model) with
fixed $z_{i}$ and $r_{i}$, it is always possible to achieve a state
in which the above condition is satisfied. 
\begin{example}
The above condition is sufficient for a solution to exist, but it
is not necessary. That is, there may be aiHRE problems, and indeed
there often are, where this condition is not met, but an algebraic
solution nevertheless exists. Consider the following case, in which
$A_{U}=\{a_{1},a_{2},a_{3}\},A_{K}=\{a_{4},a_{5}\}$ the weights of
the reference alternatives are $w(a_{4})=1.586,\,\,w(a_{5})=8.751$,
and \begingroup
\renewcommand{\arraystretch}{1.4} 
\[
C=\left[\begin{array}{ccccc}
1 & ? & 5.488 & 2.074 & ?\\
? & 1 & 1 & 5.501 & ?\\
0.182 & 1 & 1 & 6.389 & 1.119\\
0.481 & 0.181 & 0.156 & 1 & \frac{1.586}{8.751}\\
? & ? & 0.893 & \frac{8.751}{1.586} & 1
\end{array}\right].
\]
\endgroup  In this case 
\[
\overline{C}=\left[\begin{array}{ccc}
1 & 0 & -2.744\\
0 & 1 & -0.5\\
-0.0455 & -0.25 & 1
\end{array}\right],\,\,\,b=\left[\begin{array}{c}
1.644\\
4.362\\
4.981
\end{array}\right],
\]
and the solution to the equation $\overline{C}w=b$ exists and is
a valid vector $w=\left[24.129,\,8.459,\,8.194\right]^{T}$. Clearly,
$\overline{C}$ is not dominantly diagonal. 
\end{example}
Although the requirement of irreducibility postulated in Proposition
\ref{prop:A-multiplicative-incomplete} seems quite natural (we want
each pair of non-reference alternatives to be comparable), the above
requirement can be somewhat relaxed. In practice, it is sufficient
that for each alternative in each component of the graph $G(\overline{C})$,
there is one that has a direct comparison with at least one reference
alternative. Let us formulate this observation more formally. 
\begin{thm}
\label{prop:An-additive-incomplete}An arithmetic incomplete Heuristic
Rating Estimation problem for alternatives $A=A_{U_{1}}\cup\ldots\cup A_{U_{R}}\cup A_{K}$
where for every $a_{p}\in A_{U_{P}}$ and $a_{q}\in A_{U_{Q}}$ there
is no direct comparison i.e. $c_{pq}=?$ with the equation $\bar{C}w=b$
(\ref{eq:hre-addit-eq-system}) has a unique, real, and positive solution
(\ref{eq:final solution geom}) if for every $p=1,\ldots,R$ there
exists such $a_{i}\in A_{U_{p}}$ and there exists such $a_{j}\in A_{K}$
such that $c_{ij}\neq?$, the following inequality holds:\textup{
\begin{equation}
\text{\ensuremath{\left|A_{K}\right|}}-r_{i}+z_{i}\geq\sum^{k}_{\stackrel{j=1}{c_{ij}\neq?}}c_{ij},\,\,\,\text{for}\,\,\,i=1,\ldots,k.\label{eq:cond-prop}
\end{equation}
}and for every set $A_{U_{p}}$ there exists an alternative $a_{z}\in A_{U_{p}}$
such that for z-th row of $\bar{C}$ the above inequality is strict. 
\end{thm}
\begin{proof}
Because the sets of alternatives $A_{U_{1}},\ldots,A_{U_{R}}$ correspond
to connected components of a graph $G(\overline{C})$ there is a permutation
matrix $P$ such that 
\[
P\widehat{C}P^{-1}=\left[\begin{array}{ccccc}
\overline{C}_{1} & 0 & \cdots & \cdots & 0\\
0 & \overline{C}_{2} & \cdots & \cdots & 0\\
\vdots & \vdots & \overline{C}_{3} & \cdots & 0\\
\vdots & \vdots & \vdots & \ddots & 0\\
0 & \cdots & \cdots & \cdots & \overline{C}_{R}
\end{array}\right],
\]
where every $\overline{C}_{p}$ corresponds to one component of a
connected graph $G(\overline{C})$ determined by comparisons between
alternatives from $A_{U_{p}}$. Hence, the equation $\overline{C}w=b$
can be written equivalently as 
\[
\left[\begin{array}{ccccc}
\overline{C}_{1} & 0 & \cdots & \cdots & 0\\
0 & \overline{C}_{2} & \cdots & \cdots & 0\\
\vdots & \vdots & \overline{C}_{3} & \cdots & 0\\
\vdots & \vdots & \vdots & \ddots & 0\\
0 & \cdots & \cdots & \cdots & \overline{C}_{R}
\end{array}\right]\left[\begin{array}{c}
w_{1}\\
w_{2}\\
w_{2}\\
\vdots\\
w_{R}
\end{array}\right]=\left[\begin{array}{c}
\widetilde{b}_{1}\\
\widetilde{b}_{2}\\
\widetilde{b}_{3}\\
\vdots\\
\widetilde{b}_{R}
\end{array}\right],
\]

where $Pb=\widetilde{b}=\left[\widetilde{b}_{1},\widetilde{b}_{2},\widetilde{b}_{2},\ldots,\widetilde{b}_{R}\right]^{T}$.
The solution to this system is a combination of solutions to a series
of smaller systems of equations of the form $\overline{C}_{p}w_{p}=\widetilde{b}_{p}$
for $p=1,\ldots,R$. Therefore, $\overline{C}w=b$ has a solution
if each of the equations $\overline{C}_{p}w_{p}=\widetilde{b}_{p}$
has a solution. Without loss of generality, let us assume that $A_{U_{p}}=\{a_{x},\ldots,a_{y}\}$
where $1\leq x<y\leq k$. We know $\overline{C}_{p}$ is irreducible
and holds (\ref{eq:cond-prop}). Since the elements of $A_{U_{p}}$
have no comparisons with the elements of $A_{U}\backslash A_{U_{p}}$,
then (\ref{eq:cond-prop}) is true in the cut to the $x,x+1,\ldots,y$
rows and columns $\overline{C}$ i.e. for $\overline{C}_{p}$. Applying
the reasoning from the proof of Proposition \ref{prop:A-multiplicative-incomplete},
we obtain that $\widehat{C}_{p}$ is diagonally dominant. Based on
the assumption, we know that exists $a_{z}\in A_{U_{p}}$ for which
inequality (\ref{eq:cond-prop}) is sharp. Thus, the matrix $\overline{C}_{p}$
is irreducibly diagonally dominant. Therefore, due to the Corollaries
\ref{cor:postivie-matrix-observ} and \ref{cor:m-matrix-corollary},
we obtain that $\overline{C}_{p}$ is a nonsingular M-matrix where
$\overline{C}^{-1}_{p}>0$. Since, 
\[
\overline{C}^{-1}=\left[\begin{array}{ccccc}
\overline{C}^{-1}_{1} & 0 & \cdots & \cdots & 0\\
0 & \overline{C}^{-1}_{2} & \cdots & \cdots & 0\\
\vdots & \vdots & \overline{C}^{-1}_{3} & \cdots & 0\\
\vdots & \vdots & \vdots & \ddots & 0\\
0 & \cdots & \cdots & \cdots & \overline{C}^{-1}_{R}
\end{array}\right],
\]
then also $\overline{C}^{-1}$ exists and $\overline{C}^{-1}>0$.
As every $\widetilde{b}_{p}=[b_{x},\ldots,b_{y}]^{T}\geq0$, there
exists such a $b_{z}$ for $x\le z\le y$ that $b_{z}>0$ (because
for every $p=1,\ldots,R$ there exists such $a_{i}\in A_{U_{p}}$
and there exists such $a_{j}\in A_{K}$ such that $c_{ij}\neq?$),
then $\overline{C}^{-1}_{p}.\widetilde{b}_{p}=w_{p}>0$ for every
$p=1,\ldots,R$. Thus, $\overline{C}^{-1}b=w>0$.
\end{proof}

Below, let us consider an example in which $A_{U}$ consists of three
disjoint subsets $A_{U}=A_{U_{1}}\cup A_{U_{2}}\cup A_{U_{3}}$, and
each of these subsets contains one element, i.e., $A_{U_{1}}=\{a_{1}\}$,
$A_{U_{2}}=\{a_{2}\}$, and $A_{U_{3}}=\{a_{3}\}$. In other words,
in this example, no two non-referential alternatives are directly
compared with each other. However, each of these alternatives is compared
with at least one reference alternative. Thus, the model satisfies
the conditions of the Theorem \ref{prop:An-additive-incomplete} and
therefore has a solution.
\begin{example}
Let $A=A_{U}\cup A_{K}$, where $A_{U}=\{a_{1},a_{2},a_{3}\}$ and
$A_{K}=\{a_{4},a_{5}\}$, $w(a_{4})=2$ and $w(a_{5})=3$, and $C$
is in the form: 
\[
C=\left[\begin{array}{ccccc}
1 & ? & ? & c_{14} & ?\\
? & 1 & ? & c_{24} & c_{25}\\
? & ? & 1 & ? & c_{35}\\
c_{41} & c_{42} & ? & 1 & w(a_{4})/w(a_{5})\\
? & c_{52} & c_{53} & w(a_{5})/w(a_{4}) & 1
\end{array}\right].
\]

Then
\[
\overline{C}=\left[\begin{array}{ccc}
1 & 0 & 0\\
0 & 1 & 0\\
0 & 0 & 1
\end{array}\right]=\textit{Id},\,\,\,\text{and}\,\,\,b=\left[\begin{array}{c}
\frac{1}{n-s_{1}-r_{1}-1}c_{14}w(a_{4})\\
\frac{1}{n-s_{2}-r_{2}-1}\left(c_{24}w(a_{4})+c_{25}w(a_{5})\right)\\
\frac{1}{n-s_{3}-r_{3}-1}c_{35}w(a_{5})
\end{array}\right]=
\]

\[
=\left[\begin{array}{c}
\frac{1}{5-2-1-1}c_{14}w(a_{4})\\
\frac{1}{5-2-0-1}\left(c_{24}w(a_{4})+c_{25}w(a_{5})\right)\\
\frac{1}{5-2-1-1}c_{35}w(a_{5})
\end{array}\right]=\left[\begin{array}{c}
c_{14}w(a_{4})\\
\frac{c_{24}w(a_{4})+c_{25}w(a_{5})}{2}\\
c_{35}w(a_{5})
\end{array}\right],
\]

hence
\[
\left[\begin{array}{c}
w(a_{1})\\
w(a_{2})\\
w(a_{3})
\end{array}\right]=\overline{C}^{-1}b=\textit{Id}\cdot b=b
\]

As can be easily verified in the case under consideration, the weight
of the i-th non-referential alternative is the arithmetic mean of
the components of the form $c_{ij}w(a_{j})$ for which $c_{ij}\neq?$.
Obviously $\overline{C}$ is strongly dominant.
\end{example}

\subsection{The existence of a solution in the geometric approach}

Definitions \ref{def:diag-dom-matrix}, \ref{def:irreducibly-dom-matrix},
and Theorem \ref{thm:varga-theorem} allow us to formulate a condition
for the existence of a solution for a giHRE problem (\ref{eq:hre-geom-matrix-incompl}).
\begin{thm}
\label{thm:A-multiplicative-iHRE-solution-ex}A geometric incomplete
Heuristic Rating Estimation problem for alternatives $A=A_{U}\cup A_{K}$
with equation $\widehat{C}\widehat{w}=b$ (\ref{eq:hre-geom-matrix-incompl})
has a unique, real and positive solution (\ref{eq:final solution geom})
if $\widehat{C}$ is irreducible and $\exists a_{i}\in A_{U}\wedge a_{j}\in A_{K}$
for which $c_{ij}\neq?$.
\end{thm}
\begin{proof}
For the sake of convenience, let us assume that $A_{U}=a_{1},\ldots,a_{k}$
and $A_{K}=a_{k+1},\ldots,a_{n}$. Note that for each alternative
$a_{i}\in A_{U}$ the number of missing comparisons of this alternative
with the reference alternatives $a_{j}\in A_{K}$ is given as $r_{i}=\left|\{c_{ij}\,\,|\,\,c_{ij}=?\wedge1\leq i\leq k<j\leq n\}\right|$.
At the same time, the number of missing comparisons between alternative
$a_{i}$ and non-reference alternatives is denoted by $s_{i}$. The
maximum number of missing comparisons for $a_{i}\in A_{U}$ is $n-1$,
hence $n-1\geq s_{i}+r_{i}$. Similarly, the maximum number of comparisons
$a_{i}\in A_{U}$ with reference alternatives is $\left|A_{K}\right|$
i.e. $n-k\geq r_{i}$. Therefore, taking into account the criterion
(\ref{eq:diagnoally-dominance-def}) in our case, we get 

\begin{equation}
\left|c_{ii}\right|\geq\sum^{k}_{\stackrel{j=1}{j\neq i}}\left|\widehat{c}_{i,j}\right|,\label{eq:diagonally-dominant-in-theorem}
\end{equation}
hence
\[
\left|c_{ii}\right|=\left|n-s_{i}-r_{i}-1\right|=n-s_{i}-r_{i}-1\geq\sum^{k}_{\stackrel{j=1}{j\neq i}}\left|\widehat{c}_{i,j}\right|=k-s_{i}-1,
\]

what is true if and only if

\[
n-s_{i}-r_{i}-1\geq k-s_{i}-1
\]

and finally

\[
n-k\geq r_{i}.
\]
The last inequality is always true ($n-k$ is the number of reference
alternatives, $r_{i}$ - number of missing comparisons between the
i-th alternative and reference alternatives), which makes the matrix
$\widehat{C}$ diagonally dominant. However, based on the assumptions,
we know that for at least one pair of alternatives $a_{i}\in A_{U}$
and $a_{j}\in A_{K}$, there is a comparison $c_{ij}\neq?$. This
means that for $a_{i}$, the number of missing comparisons with reference
alternatives is no greater than $r_{i}=n-k-1$. Thus, the inequality
in (\ref{eq:diagonally-dominant-in-theorem}) is sharp. Since the
matrix $\widehat{C}$ is assumed to be irreducible, $\widehat{C}$
is irreducibly diagonally dominant, which, based on Theorem \ref{thm:varga-theorem},
means that $\widehat{C}$ is nonsingular. Thus, the equation (\ref{eq:hre-geom-matrix-incompl})
has a unique and real solution, i.e. the weight vector (\ref{eq:final solution geom})
for the original problem, it is always unique, real, and positive. 
\end{proof}

The conditions formulated above for the existence of a solution to
the incomplete HRE problem are, in fact, quite natural. The requirement
that the matrix $\widehat{C}$ be irreducible means that all non-referential
alternatives must be directly or indirectly comparable to each other.
A similar requirement exists for an incomplete pairwise comparison
matrix in the EVM \citep{Harker1987amoq} or GMM \citep{kulakowski2020otgm}
approach. It is not easy to require the construction of a common weight
vector for alternatives that cannot be compared with each other. Similarly,
the requirement that there be a pair of alternatives $a_{i}\in A_{U}$
and $a_{k}\in A_{K}$ with a direct comparison $c_{ij}$ means precisely
that in the set of all alternatives $A=A_{U}\cup A_{K}$, any two
are directly or indirectly comparable.

It is worth noting that Theorem \ref{thm:A-multiplicative-iHRE-solution-ex}
guarantees that when the matrix has a strongly dominant diagonal,
i.e., inequality (\ref{eq:diagnoally-dominance-def}) is strict, irreducibility
is not necessary for a matrix to be nonsingular. This leads to the
following immediate observation.
\begin{cor}
A geometric incomplete Heuristic Rating Estimation problem for alternatives
$A=A_{U}\cup A_{K}$ with equation $\widehat{C}\widehat{w}=b$ (\ref{eq:hre-geom-matrix-incompl})
has a unique, real, and positive solution (\ref{eq:final solution geom})
if $\forall a_{i}\in A_{U}\exists a_{j}\in A_{K}$ such that $c_{ij}\neq?$.
\end{cor}
This means that we can successfully replace the condition of irreducibility
with the requirement that each non-reference alternative be compared
with at least one reference alternative. In practice, we can relax
this requirement and demand that in each component of a connected
graph $G(\widehat{C})$ associated with the matrix $\widehat{C}$
there is at least one alternative directly comparable to at least
one reference alternative.
\begin{thm}
\label{thm:A-multiplicative-iHRE-solution-ex-general}A geometric
incomplete Heuristic Rating Estimation problem for alternatives $A=A_{U_{1}}\cup\ldots A_{U_{R}}\cup A_{K}$,
where for every $a_{p}\in A_{U_{P}}$ and $a_{q}\in A_{U_{Q}}$ there
is no direct comparison, i.e., $c_{pq}=?$ the equation $\widehat{C}\widehat{w}=b$
(\ref{eq:hre-geom-matrix-incompl}) has a unique, real, and positive
solution (\ref{eq:final solution geom}) if for every $A_{U_{I}}$
there exists $a_{i}\in A_{U_{I}}$ and $a_{j}\in A_{K}$ with direct
comparison, i.e., $c_{ij}\neq?$. 
\end{thm}
\begin{proof}
Because the sets of alternatives $A_{U_{1}},\ldots,A_{U_{R}}$ determine
the connected components of a graph $G(\widehat{C})$ there is a permutation
matrix $P$ such that 
\[
P\widehat{C}P^{-1}=\left[\begin{array}{ccccc}
\widehat{C}_{1} & 0 & \cdots & \cdots & 0\\
0 & \widehat{C}_{2} & \cdots & \cdots & 0\\
\vdots & \vdots & \widehat{C}_{3} & \cdots & 0\\
\vdots & \vdots & \vdots & \ddots & 0\\
0 & \cdots & \cdots & \cdots & \widehat{C}_{R}
\end{array}\right],
\]
where every $\widehat{C}_{p}$ corresponds to one component of a connected
graph $G(\widehat{C})$ determined by comparisons between alternatives
in the set $A_{U_{p}}$, hence, the equation $\widehat{C}\widehat{w}=b$
can be written equivalently as 
\[
\left[\begin{array}{ccccc}
\widehat{C}_{1} & 0 & \cdots & \cdots & 0\\
0 & \widehat{C}_{2} & \cdots & \cdots & 0\\
\vdots & \vdots & \widehat{C}_{3} & \cdots & 0\\
\vdots & \vdots & \vdots & \ddots & 0\\
0 & \cdots & \cdots & \cdots & \widehat{C}_{R}
\end{array}\right]\left[\begin{array}{c}
\widehat{w}_{1}\\
\widehat{w}_{2}\\
\widehat{w}_{2}\\
\vdots\\
\widehat{w}_{R}
\end{array}\right]=\left[\begin{array}{c}
\widetilde{b}_{1}\\
\widetilde{b}_{2}\\
\widetilde{b}_{3}\\
\vdots\\
\widetilde{b}_{R}
\end{array}\right],
\]

where $Pb=\widetilde{b}=\left[\widetilde{b}_{1},\widetilde{b}_{2},\widetilde{b}_{2},\ldots,\widetilde{b}_{R}\right]^{T}$.
The solution to this system is a combination of solutions to a series
of smaller systems of equations of the form $\widehat{C}_{p}\widehat{w}_{p}=\widetilde{b}_{p}$
for $p=1,\ldots,R$. Each matrix $\widehat{C}_{p}$ is irreducible,
and we also know that it is diagonally dominant (for reasons presented
in the proof of Theorem \ref{thm:A-multiplicative-iHRE-solution-ex}).
Furthermore, since there is at least one alternative $a_{p}\in\widehat{C}_{p}$
directly comparable to some other reference alternative $a_{q}\in A_{K}$,
i.e., $c_{pq}\neq?$, then, similarly to Theorem \ref{thm:A-multiplicative-iHRE-solution-ex},
by Varga's (Theorem \ref{thm:varga-theorem}), $\widehat{C}_{p}$
is invertible. Since every $\widetilde{b}_{p}>0$, then $\widehat{C}^{-1}_{p}\widetilde{b}_{p}\neq0$.
Thus $\widehat{w}\neq0$, and therefore $w>0$. 
\end{proof}

\begin{example}
Let us consider an example where $A=A_{U_{1}}\cup A_{U_{2}}\cup A_{K}$,
where $A_{U_{1}}=\{a_{1},a_{2},a_{3}\}$, $A_{U_{2}}=\{a_{4},a_{5},a_{6}\}$
and $A_{K}=\{a_{7},a_{8}\}$. The reference values for alternatives
from $A_{K}$ are as follows $w(a_{7})=5$ and $w(a_{8})=9$. As a
result of the expert's work, we obtained the following comparisons
matrix: 

\begingroup
\renewcommand{\arraystretch}{1.4}
\[
C=\left[\begin{array}{cccccccc}
1 & 2 & ? & ? & ? & ? & 2 & ?\\
\frac{1}{2} & 1 & 3 & ? & ? & ? & ? & ?\\
? & \frac{1}{3} & 1 & ? & ? & ? & ? & ?\\
? & ? & ? & 1 & \frac{1}{3} & 2 & ? & ?\\
? & ? & ? & 3 & 1 & ? & ? & \frac{3}{2}\\
? & ? & ? & \frac{1}{2} & ? & 1 & ? & ?\\
\frac{1}{2} & ? & ? & ? & ? & ? & 1 & \frac{5}{9}\\
? & ? & ? & ? & \frac{2}{3} & ? & \frac{9}{5} & 1
\end{array}\right].
\]

\endgroup  The system of equations matrix and constant vector for
a giHRE are 
\[
\widehat{C}=\left[\begin{array}{cccccc}
2 & -1 & 0 & 0 & 0 & 0\\
-1 & 2 & -1 & 0 & 0 & 0\\
0 & -1 & 1 & 0 & 0 & 0\\
0 & 0 & 0 & 2 & -1 & -1\\
0 & 0 & 0 & -1 & 2 & 0\\
0 & 0 & 0 & -1 & 0 & 1
\end{array}\right],\,\,\,b=\left[\begin{array}{c}
\log(2)+\log(10)\\
\log(3)-\log(2)\\
-\log(3)\\
\log(2)-\log(3)\\
\log(3)+\log\left(\frac{27}{2}\right)\\
-\log(2)
\end{array}\right]=\left[\begin{array}{c}
2.995\\
0.405\\
-1.098\\
-0.405\\
3.701\\
-0.693
\end{array}\right].
\]

Therefore, the solution to the linear problem is
\[
\widehat{C}^{-1}b=\left[\begin{array}{c}
\log(10)\\
\log(2)-2\log(3)+2(\log(3)-\log(2))+\log(10)\\
\log(2)-3\log(3)+2(\log(3)-\log(2))+\log(10)\\
-2\log(2)+2(\log(2)-\log(3))+\log(3)+\log\left(\frac{27}{2}\right)\\
\log\left(\frac{27}{2}\right)\\
-3\log(2)+2(\log(2)-\log(3))+\log(3)+\log\left(\frac{27}{2}\right)
\end{array}\right]=\left[\begin{array}{c}
2.302\\
1.609\\
0.51\\
1.504\\
2.602\\
0.81
\end{array}\right]=\widehat{w}.
\]

Finally, $w=\exp(\widehat{w})=\left[10,5,\frac{5}{3},\frac{9}{2},\frac{27}{2},\frac{9}{4}\right]^{T}$,
and after normalization $w/\left\Vert w\right\Vert _{1}=\left[0.27,0.13,0.045,0.121,0.365,0.06\right]^{T}$. 

It is worth noting that we would obtain the same result by solving
two smaller systems of equations $\widehat{C}_{1}\widehat{w}_{1}=\widetilde{b}_{1}$
and $\widehat{C}_{1}\widehat{w}_{2}=\widetilde{b}_{2}$, separately,
where
\[
\widehat{C}_{1}=\left[\begin{array}{ccc}
2 & -1 & 0\\
-1 & 2 & -1\\
0 & -1 & 1
\end{array}\right],\,\,\widetilde{b}_{1}=\left[\begin{array}{c}
\log(2)+\log(10)\\
\log(3)-\log(2)\\
-\log(3)
\end{array}\right],
\]
and

\[
\widehat{C}_{2}=\left[\begin{array}{ccc}
2 & -1 & -1\\
-1 & 2 & 0\\
-1 & 0 & 1
\end{array}\right],\,\,\widetilde{b}_{2}=\left[\begin{array}{c}
\log(2)-\log(3)\\
\log(3)+\log\left(\frac{27}{2}\right)\\
-\log(2)
\end{array}\right].
\]
\end{example}

\subsection{Optimality of the geometric approach}

In the original GMM method proposed by Crawford \citep{Crawford1987tgmp},
the geometric method minimizes the total error function, defined as
the sum of the logarithms of the products of direct comparisons and
their corresponding weight ratios. Indeed, if we assume that the error
for a single comparison is
\[
\epsilon_{ij}=\ln\left(c_{ij}\frac{w(a_{j})}{w(a_{i})}\right),
\]
then it is zero if and only if $c_{ij}=w(a_{i})/w(a_{j})$. The total
error for the pairwise comparison matrix $C=[c_{ij}]$ is given by
the function 
\[
\mathcal{E}(C,w)=\sum^{n}_{i,j=1}\epsilon^{2}_{ij}=\sum^{n}_{i,j=1}\left(\ln c_{ij}-\ln\left(\frac{w(a_{j})}{w(a_{i})}\right)\right)^{2}=\left\Vert \ln C-\ln\left(w\cdot w^{-T}\right)\right\Vert ^{2}_{2},
\]
where $w^{-T}$ is a vector $w=\left[w^{-1}(a_{1}),\ldots,w^{-1}(a_{n})\right]^{T}$
and $\left\Vert \cdot\right\Vert _{2}$ is the Frobenius norm. The
function $\mathcal{E}(C,w)$ reaches its minimum \citep{Crawford1987tgmp},
i.e.
\[
\frac{\partial\mathcal{E}(C,w)}{\partial w(a_{r})}=0\,\,\text{for all}\,\,r=1,\ldots,n
\]
when 
\begin{equation}
w(a_{i})=\left(\prod^{n}_{j=1}c_{ij}w(a_{j})\right)^{1/n}\,\,\text{for all}\,\,i=1,\ldots,n,\label{eq:gmm-assumption}
\end{equation}
which is equivalent to assuming that the weights of the individual
alternatives are the geometric means of the rows of the matrix $C=[c_{ij}]$
, i.e. 
\[
w(a_{i})=\left(\prod^{n}_{j=1}c_{ij}\right)^{1/n}\,\,\text{for all}\,\,i=1,\ldots,n.
\]
For convenience, these values are normalized so that their sum equals
one (see \ref{eq:GMM-eq}). 

Let us define an analogous total error function for an incomplete
matrix $C$ (\ref{eq:incomplete-matrix-in-hre}) in giHRE. 
\[
\mathcal{E}^{*}(C,w)=\sum^{n}_{\substack{i,j=1\\
c_{ij}\neq?\\
a_{i}\notin A_{K}\vee a_{j}\notin A_{K}
}
}\left(\ln c_{ij}-\ln\left(\frac{w(a_{j})}{w(a_{i})}\right)\right)^{2}.
\]

That is, we want $\ln\epsilon_{ij}$ to be as close to zero as possible,
but only in those cases where a direct comparison exists, and the
alternatives being compared are not reference alternatives. 
\begin{prop}
The solution of the geometric incomplete HRE is optimal, i.e., the
result obtained minimizes the value of the error function $\mathcal{E}^{*}(C)$.
\end{prop}
\begin{proof}
As before, without loss of generality, let us assume that $A_{U}=\{a_{1},\ldots,a_{k}\}$
and $A_{K}=\{a_{k+1},\ldots,a_{n}\}$. Since giHRE is essentially
the GMM procedure applied to the $C^{*}$ matrix (\ref{eq:complemented-matrix}),
assuming that the weights for alternatives from $A_{K}$ are defined,
we can observe that
\[
\mathcal{E}(C^{*},w)=\sum^{n}_{\substack{i,j=1}
}\left(\ln c^{*}_{ij}-\ln\left(\frac{w(a_{j})}{w(a_{i})}\right)\right)^{2}=
\]

\[
\sum^{n}_{\substack{i,j=1\\
c_{ij}\neq?
}
}\left(\ln c_{ij}-\ln\left(\frac{w(a_{j})}{w(a_{i})}\right)\right)^{2}+\underset{0}{\underbrace{\sum^{n}_{\substack{i,j=1\\
c_{ij}=?
}
}\left(\ln\left(\frac{w(a_{j})}{w(a_{i})}\right)-\ln\left(\frac{w(a_{j})}{w(a_{i})}\right)\right)^{2}}}=
\]
\[
=\sum^{k}_{\substack{i,j=1\\
c_{ij}\neq?
}
}\left(\ln c_{ij}-\ln\left(\frac{w(a_{j})}{w(a_{i})}\right)\right)^{2}+\sum_{\substack{i=k+1,\ldots,n\\
j=1,\ldots,k\\
c_{ij}\neq?
}
}\left(\ln c_{ij}-\ln\left(\frac{w(a_{j})}{w(a_{i})}\right)\right)^{2}+
\]

\[
+\sum_{\substack{i=1,\ldots,k\\
j=k+1,\ldots,n\\
c_{ij}\neq?
}
}\left(\ln c_{ij}-\ln\left(\frac{w(a_{j})}{w(a_{i})}\right)\right)^{2}=
\]
\[
=\sum^{n}_{\substack{i,j=1\\
c_{ij}\neq?\\
a_{i}\notin A_{K}\vee a_{j}\notin A_{K}
}
}\left(\ln c_{ij}-\ln\left(\frac{w(a_{j})}{w(a_{i})}\right)\right)^{2}=\mathcal{E}^{*}(C,w).
\]

Since $\mathcal{E}(C^{*},w)=\mathcal{E}^{*}(C,w)$ and we know that
$\mathcal{E}(C^{*},w)$ reaches a minimum on a complete (supplemented)
matrix $C^{*}$ with the assumption (\ref{eq:gmm-assumption}) (see
\ref{eq:geom-nonlinear-eq-system}), then also $\mathcal{E}^{*}$
achieves a minimum for an incomplete matrix $C$ (\ref{eq:incomplete-matrix-in-hre})
where alternatives $a_{k+1},\ldots,a_{n}$ have their values $w(a_{k+1}),\ldots,w(a_{n})$
known and fixed.
\end{proof}

\section{Discussion\label{sec:Discussion}}

A prevalent question in the context of the AHP method is: which method
for calculating the weight vector should be chosen? The two most commonly
used methods, EVM and GMM, have their opponents and supporters who
rightly argue for one or the other. One can read about several primary
methods, including EVM and GMM, for example, in \citep{Srdjevic2023pita}.
Similarly, in the case of arithmetic and geometric HRE, we can ask
a similar question. As long as the pairwise comparison matrix $C=[c_{ij}]$
is consistent, i.e., $c_{ij}=c_{ik}c_{kj}$ for $1\leq i,k,j\leq n$,
the answer is straightforward. Both methods return the same result.
To see this, let us first show that $w(a_{i})/w(a_{r})=c_{ir}$. More
specifically, we will show this fact for the geometric approach, since
the reasoning in the arithmetic case is the same, with the accuracy
of the mean used. 
\begin{prop}
\label{prop:For-a-consistent}For a consistent (but not necessarily
complete) pairwise comparison matrix $C=[c_{ij}]$ for any two indices
$i$ and $j$, it holds $w(a_{i})=c_{ij}w(a_{j})$, where either $c_{ij}$
is either directly specified or uniquely inferable. 
\end{prop}
\begin{proof}
Due to the incompleteness of matrix $C$, either $c_{ir}$ or one
of $c_{ik}$ may be undefined. However, since $C$ is irreducible,
for every matrix entry of the form $c_{pq}$, there exists a sequence
$S_{pq}$ of the form $S_{pq}=\left(c_{p,s_{1}},c_{s_{1},s_{2}},\ ...c_{s_{x},q}\right)$
consisting of values directly defined in $C$. Let us denote the product
of elements on the path $S_{pq}$ as $\pi_{pq}$. Since $C$ is consistent,
for any element $c_{pq}$, it holds that $c_{pq}=\pi_{pq}$. In other
words, if $c_{pq}\neq?$, we can use this value directly. However,
if $c_{pq}=?$, instead of using it directly, we can take the value
$\pi_{pq}$, which, due to the consistency of $C$, equals $c_{pq}$.
Let us assume
\[
\overline{c}_{pq}=\begin{cases}
c_{pq} & \text{if}\,\,c_{pq}\neq?\\
\pi_{pq} & \text{if}\,\,c_{pq}=?
\end{cases}.
\]
Let us consider the equation (\ref{eq:incompl-geometric-hre-eq})
that underlies the giHRE method. After raising both sides to the power
of $n-1$, we obtain: 
\[
w^{n-1}(a_{i})=w^{s_{i}+r_{i}}(a_{i})\underset{n-s_{i}-r_{i}-1}{\underbrace{\prod^{n}_{\substack{k=1,i\neq j\\
c_{ik}\neq?
}
}c_{ik}w(a_{k})}},
\]
and further
\[
w^{n-1}(a_{i})=\prod^{n}_{\substack{k=1\\
i\neq j
}
}\overline{c}_{ik}w(a_{k}).
\]

Consider the quotient 
\[
\frac{w^{n-1}(a_{i})}{w^{n-1}(a_{j})}=\prod^{n}_{\substack{k=1\\
i\neq k
}
}\overline{c}_{ik}w(a_{k})\left/\prod^{n}_{\substack{k=1\\
j\neq k
}
}\overline{c}_{jk}w(a_{k})\right..
\]

Continuing the transformation 
\[
\frac{w^{n-1}(a_{i})}{w^{n-1}(a_{j})}=\prod^{n}_{\substack{k=1\\
i\neq k
}
}\overline{c}_{ir}\overline{c}_{rk}w(a_{k})\left/\prod^{n}_{\substack{k=1\\
j\neq k
}
}\overline{c}_{jr}\overline{c}_{rk}w(a_{k}),\right.
\]
we get 
\[
\frac{w^{n-1}(a_{i})}{w^{n-1}(a_{j})}=\left(\frac{\overline{c}_{ir}}{\overline{c}_{jr}}\right)^{n-1}\underset{=1}{\underbrace{\left(\prod^{n}_{\substack{k=1\\
i\neq k
}
}\overline{c}_{rk}w(a_{k})\left/\prod^{n}_{\substack{k=1\\
j\neq k
}
}\overline{c}_{rk}w(a_{k})\right.\right)}},
\]
and finally
\[
\frac{w(a_{i})}{w(a_{j})}=\frac{\overline{c}_{ir}}{\overline{c}_{jr}}=\overline{c}_{ir}\overline{c}_{rj}=\overline{c}_{ij}.
\]

Hence the desired conclusion: 
\[
w(a_{i})=\overline{c}_{ij}w(a_{j}).
\]
\end{proof}

The reasoning behind the above proposal for aiHRE is analogous. Thanks
to the above Proposition \ref{prop:For-a-consistent}, it is easy
to show that in the case of a consistent pairwise comparison matrix,
both methods, i.e., aiHRE and giHRE, lead to the same weight vector.
\begin{prop}
For a consistent PC matrix $C=[c_{ij}]$, the priority values assigned
to non-reference alternatives by aiHRE and giHRE are identical.
\end{prop}
\begin{proof}
Let $a_{r}\in A_{U}$ be the chosen reference alternative. Let us
assume $w_{\textit{aihre}}(a_{r})=w_{\textit{gihre}}(a_{r})=\alpha\in\mathbb{R}_{+}$,
where $w_{\textit{aihre}}$ are the priority values in the aiHRE approach
and $w_{\textit{gihre}}$ are the priority values in the giHRE approach.
Therefore, for any non-reference alternative $a_{t}\in A_{U}$, we
have
\[
w_{\textit{aihre}}(a_{t})=\overline{c}_{tr}w_{\textit{aihre}}(a_{r})=\alpha=\overline{c}_{tr}w_{\textit{gihre}}(a_{r})=w_{\textit{gihre}}(a_{t}).
\]
\end{proof}

The answer to the question of which method to choose begins to matter
for inconsistent matrices. There are compelling arguments for using
both methods. The geometric approach is supported by the existence
of a solution in virtually every situation and by the solution's optimality.
The arithmetic approach is supported by the arithmetic mean property,
which has a reasonable and intuitive interpretation. The existence
of reference values gives these arguments additional practical significance.

\begin{example}
Let us consider the following example, in which the set of alternatives
consists of three non-reference alternatives $A_{U}=\{a_{1},a_{2},a_{3}\}$,
and two reference alternatives $A_{K}=\{a_{4},a_{5}\}$ where the
values of the latter are $\$3$ and $\$5$, respectively. For the
sake of attention, let us assume that the alternatives are types of
souvenirs in a souvenir shop, the reference values are the prices
of the souvenirs in US dollars, and the goal of the process is to
determine the initial prices of the products $a_{1}$, $a_{2}$, and
$a_{3}$. When making comparisons, the DM attempts to estimate the
values of alternatives $A_{U}$ relative to $A_{K}$, and the pairwise
comparison matrix looks as follows: \begingroup
\renewcommand{\arraystretch}{1.4}

\[
C=\left[\begin{array}{ccccc}
1 & ? & ? & 2 & 4\\
? & 1 & ? & 3 & \frac{1}{2}\\
? & ? & 1 & 3 & 5\\
\frac{1}{2} & \frac{1}{3} & \frac{1}{3} & 1 & \frac{3}{5}\\
\frac{1}{4} & 2 & 5 & \frac{5}{3} & 1
\end{array}\right].
\]

\endgroup  When calculating results using aiHRE, the resulting weight
vector is:
\[
w_{\text{\textit{aihre}}}=\left[13,5.75,17\right]^{T},
\]
whereas the geometric method returns:

\[
w_{\textit{gihre}}=\left[10.954,\,4.743,\,15\right]^{T}.
\]
The result of the arithmetic method is easy to interpret and, in this
case, fairly simple to calculate in mind. For example, $w_{\textit{aihre}}(a_{2})=\frac{1}{2}\left(3\cdot3+\frac{1}{2}5\right)=5.75$.
The DM decided that product $a_{2}$ should be three times more expensive
than product $a_{4}$ ($3\cdot3=9$) and, at the same time, half as
expensive as product $a_{5}$, i.e., $\frac{1}{2}\cdot5=2.5$. Since
both of these judgments are a priori equally valid (equally probable),
the result is the arithmetic mean of the estimates. On this basis,
DM concludes that product $a_{2}$ should initially cost $\$5.75$.

In the geometric case, the obtained value $w_{\textit{gihre}}(a_{2})=(3\cdot3\cdot\frac{1}{2}\cdot5)^{1/2}=4.74$
is obviously smaller than $w_{\textit{aihre}}(a_{2})$. The result
is definitely influenced by the judgment that the value $a_{2}$ that
is half the value of $a_{5}$. Both comparisons are not treated equally,
and the more pessimistic estimate has a greater impact on the final
result. The question arises whether, based on the value $w_{\textit{gihre}}(a_{2})$,
we should offer a lower price and accept a lower margin, or choose
a higher price $w_{\textit{aihre}}(a_{2})$? 
\end{example}
In the above example, our conviction that both of the judgments, i.e.,
the comparison of $a{{}_2}$ with $a{{}_4}$ and $a{{}_2}$ with $a{{}_5}$,
are equally valid, speaks in favor of $w_{\textit{aihre}}(a_{2})$.
On the other hand, theoretical properties (existence of a solution,
optimality) speak in favor of $w_{\textit{gihre}}(a_{2})$. The question
of which method to choose, therefore, is a question of whether one
argument seems more relevant in a given situation than the other. Which
problem are we dealing with? Is this an attempt to estimate specific
values, assuming that each assessment is equally important, or is
it a calculation of weight vectors for ranking? In the first case,
the arithmetic approach may be more appropriate. In the second case,
we believe that the advantages of the geometric approach may become
apparent. Ultimately, in each case, the decision-maker will have to
choose which method to use. 

\section{Summary\label{sec:Summary}}

In this paper, we introduced two variants - arithmetic and geometric
- of a quantitative pairwise comparison method that derives a weight
vector from incomplete pairwise comparison matrices while incorporating
reference alternatives. The proposed methods substantially reduce
the effort required to conduct the decision-making process and, consequently,
lower the cost of generating recommendations. We established conditions
for the existence of solutions in both decision problems and proved
the optimality of the computed weight vector in the geometric case.
 The proposed HRE methods for incomplete pairwise comparison matrices
integrate naturally with the hierarchical structure known from the
AHP method, increasing its flexibility by enabling the use of reference
values and reducing the number of pairwise comparisons at each node
of the model tree. Given the advantages of the proposed approach and
its compatibility with AHP, we believe that the presented solutions
may be of interest to both practitioners and researchers working in
the field of multi-criteria decision-making.

\section*{Acknowledgments}

The research has been supported by the National Science Centre, Poland
within the grant VIRGO 2024/55/B/HS4/00860.

\bibliographystyle{elsarticle-harv}
\addcontentsline{toc}{section}{\refname}\bibliography{../EJOR_1/papers_biblio_reviewed}

\end{document}